\documentclass[11pt,a4paper]{article}
\usepackage{times,latexsym}
\usepackage{url}
\usepackage[T1]{fontenc}

\usepackage[acceptedWithA]{tacl2018v2}
\usepackage{graphicx}
\usepackage{amsmath}
\usepackage{amssymb}
\usepackage{caption}
\usepackage{algorithm}
\usepackage{algorithmicx}
\usepackage{algpseudocode}
\usepackage{multicol}
\usepackage{multirow}
\usepackage{booktabs}
\usepackage{adjustbox}
\usepackage{listings}
\usepackage{url}
\usepackage{bm}
\usepackage{xspace}
\usepackage{makecell}
\usepackage{lipsum}
\usepackage{comment}
\usepackage{subcaption}

\usepackage{microtype}

\usepackage{xspace,mfirstuc,tabulary}

\newif\iftaclinstructions
\taclinstructionsfalse 
\iftaclinstructions
\renewcommand{\confidential}{}
\renewcommand{\anonsubtext}{(No author info supplied here, for consistency with
TACL-submission anonymization requirements)}
\newcommand{\instr}
\fi

\newcommand\tablefont{
    \small
}

\title{KEPLER: A Unified Model for Knowledge Embedding and \\ Pre-trained Language Representation}

\author{Xiaozhi Wang$^1$, Tianyu Gao$^3$, Zhaocheng Zhu$^{4,5}$, Zhengyan Zhang$^1$, \\ \textbf{Zhiyuan Liu}$^{1,2*}$, \textbf{Juanzi Li}$^{1,2}$, \textbf{Jian Tang}$^{4,6,7}$\thanks{\quad Correspondence to: Z.~Liu and J.~Tang}\hspace{0.5em}\\
$^{1}$Department of CST, BNRist;
$^{2}$KIRC, Institute for AI, Tsinghua University, Beijing, China\\
\texttt{\{wangxz20,zy-z19\}@mails.tsinghua.edu.cn} \\ \texttt{\{liuzy,lijuanzi\}@tsinghua.edu.cn} \\
$^3$Department of Computer Science, Princeton University, Princeton, USA\\
\texttt{tianyug@princeton.edu}\\ 
$^4$Mila - Qu\'ebec AI Institute; $^5$Univesit\'e de Montr\'eal; $^6$HEC, Montr\'eal, Canada\\
\texttt{zhaocheng.zhu@umontreal.ca}, \texttt{jian.tang@hec.ca}\\
 $^7$CIFAR AI Research Chair\\
}

\date{}

\begin{document}
\maketitle
\begin{abstract}

Pre-trained language representation models (PLMs) cannot well capture factual knowledge from text. In contrast, knowledge embedding (KE) methods can effectively represent the relational facts in knowledge graphs (KGs) with informative entity embeddings, but conventional KE models cannot take full advantage of the abundant textual information.
In this paper, we propose a unified model for \textbf{K}nowledge \textbf{E}mbedding and \textbf{P}re-trained \textbf{L}anguag\textbf{E} \textbf{R}epresentation (\textbf{KEPLER}), which can not only better integrate factual knowledge into PLMs but also produce effective text-enhanced KE with the strong PLMs. 
In KEPLER, we encode textual entity descriptions with a PLM as their embeddings, and then jointly optimize the KE and language modeling objectives. 
Experimental results show that KEPLER achieves state-of-the-art performances on various NLP tasks, and also works remarkably well as an inductive KE model on KG link prediction.
Furthermore, for pre-training and evaluating KEPLER, we construct Wikidata5M\footnote{\url{https://deepgraphlearning.github.io/project/wikidata5m}}, a large-scale KG dataset with aligned entity descriptions, and benchmark state-of-the-art KE methods on it. It shall serve as a new KE benchmark and facilitate the research on large KG, inductive KE, and KG with text. The source code can be obtained from~\url{https://github.com/THU-KEG/KEPLER}.

\end{abstract}

\section{Introduction}

Recent pre-trained language representation models (PLMs) such as BERT~\citep{devlin-etal-2019-bert} and RoBERTa~\citep{liu2019roberta} learn effective language representation from large-scale unstructured corpora with language modeling objectives and have achieved superior performances on various natural language processing (NLP) tasks. 
Existing PLMs learn useful linguistic knowledge from unlabeled text~\citep{liu-etal-2019-linguistic}, but they generally cannot well capture the world facts, which are typically sparse and have complex forms in text~\citep{petroni2019language,logan-etal-2019-baracks}.
\begin{figure}[t]
    \centering
    \includegraphics[width=0.98\linewidth]{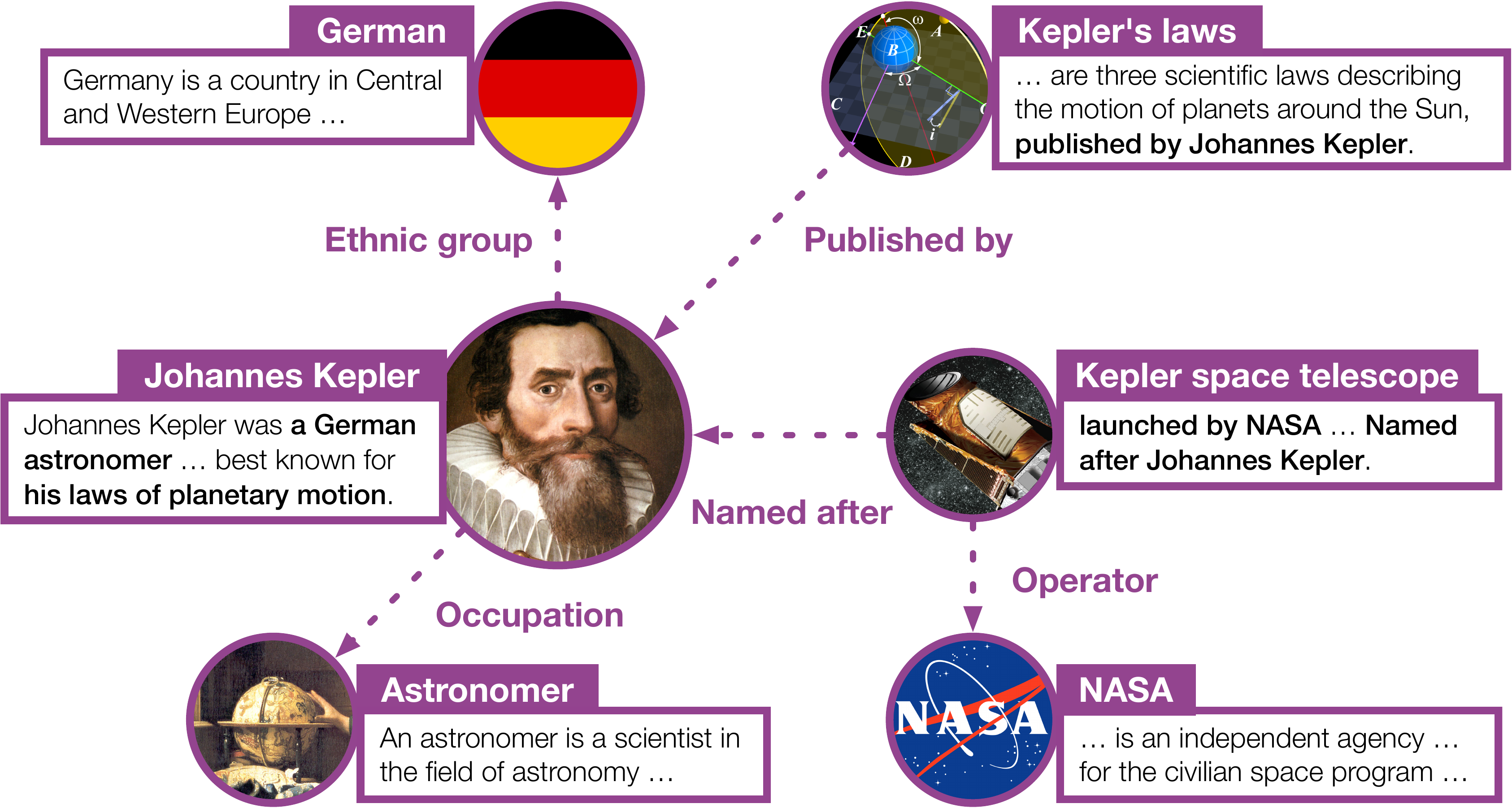}
    \caption{An example of a KG with entity descriptions. The figure suggests that descriptions contain abundant information about entities and can help to represent the relational facts between them.}
    \label{fig:kepler_head}
\end{figure}

By contrast, knowledge graphs (KGs) contain extensive structural facts, and knowledge embedding (KE) methods~\citep{bordes2013translating, yang2015embedding, sun2019rotate} can effectively embed them into continuous entity and relation embeddings. These embeddings can not only help with the KG completion but also benefit various NLP applications~\citep{yang-mitchell-2017-leveraging,zaremoodi-etal-2018-adaptive, han2018neural}. As shown in Figure~\ref{fig:kepler_head}, textual entity descriptions contain abundant information. Intuitively, KE methods can provide factual knowledge for PLMs, while the informative text data can also benefit KE.

Inspired by \citet{Xie:2016:RLK:3016100.3016273}, we use entity descriptions to bridge the gap between KE and PLM, and align the semantic space of text to the symbol space of KGs~\citep{logeswaran-etal-2019-zero}. We propose \textbf{KEPLER}, a unified model for \textbf{K}nowledge \textbf{E}mbedding and \textbf{P}re-trained \textbf{L}anguag\textbf{E} \textbf{R}epresentation. We encode the texts and entities into a unified semantic space with the same PLM as the encoder, and jointly optimize the KE and the masked language modeling (MLM) objectives. For the KE objective, we encode the entity descriptions as entity embeddings and then train them in the same way as conventional KE methods. For the MLM objective, we follow the approach of existing PLMs~\cite{devlin-etal-2019-bert,liu2019roberta}. KEPLER has the following strengths:

\textbf{As a PLM}, (1) KEPLER is able to integrate factual knowledge into language representation with the supervision from KG by the KE objective. (2) KEPLER inherits the strong ability of language understanding from PLMs by the MLM objective. (3) The KE objective enhances the ability of KEPLER to extract knowledge from text since it requires the model to encode the entities from their corresponding descriptions. (4) KEPLER can be directly adopted in a wide range of NLP tasks without additional inference overhead compared to conventional PLMs since we just add new training objectives without modifying model structures.

There are also some recent works~\citep{zhang-etal-2019-ernie,peters-etal-2019-knowledge,liu2019k} directly integrating fixed entity embeddings into PLMs to provide external knowledge. However, (1) their entity embeddings are learned by a separate KE model, and thus cannot be easily aligned with the language representation space. (2) They require an entity linker to link the text to the corresponding entities, making them suffer from the error propagation problem. (3) Compared to vanilla PLMs, their sophisticated mechanisms to link and use entity embeddings lead to additional inference overhead.

\textbf{As a KE model}, (1) KEPLER can take full advantage of the abundant information from entity descriptions with the help of the MLM objective. (2) KEPLER is capable of performing KE in the inductive setting, i.e., it can produce embeddings for unseen entities from their descriptions, while conventional KE methods are inherently transductive and they can only learn representations for the shown entities during training. Inductive KE is essential for many real-world applications, such as updating KGs with emerging entities and KG construction, and thus is worth more investigation.

For pre-training and evaluating KEPLER, we need a KG with (1) large amounts of knowledge facts, (2) aligned entity descriptions, and (3) reasonable inductive-setting data split, which cannot be satisfied by existing KE benchmarks. Therefore, we construct Wikidata5M, containing about 5M entities, 20M triplets, and aligned entity descriptions from Wikipedia. To the best of our knowledge, it is the largest general-domain KG dataset. We also benchmark several classical KE methods and give data splits for both the transductive and the inductive settings to facilitate future research. 

To summarize, our contribution is three-fold: \linebreak
(1) We propose KEPLER, a knowledge-enhanced PLM by jointly optimizing the KE and MLM objectives, which brings great improvements on a wide range of NLP tasks. (2) By encoding text descriptions as entity embeddings, KEPLER shows its effectiveness as a KE model, especially in the inductive setting. (3) We also introduce Wikidata5M, a new large-scale KG dataset, which shall promote the research on large-scale KG, inductive KE, and the interactions between KG and NLP.
\section{KEPLER}
\label{sec:kepler}
 \begin{figure*}[t]
\centering
\includegraphics[width = 0.98\textwidth]{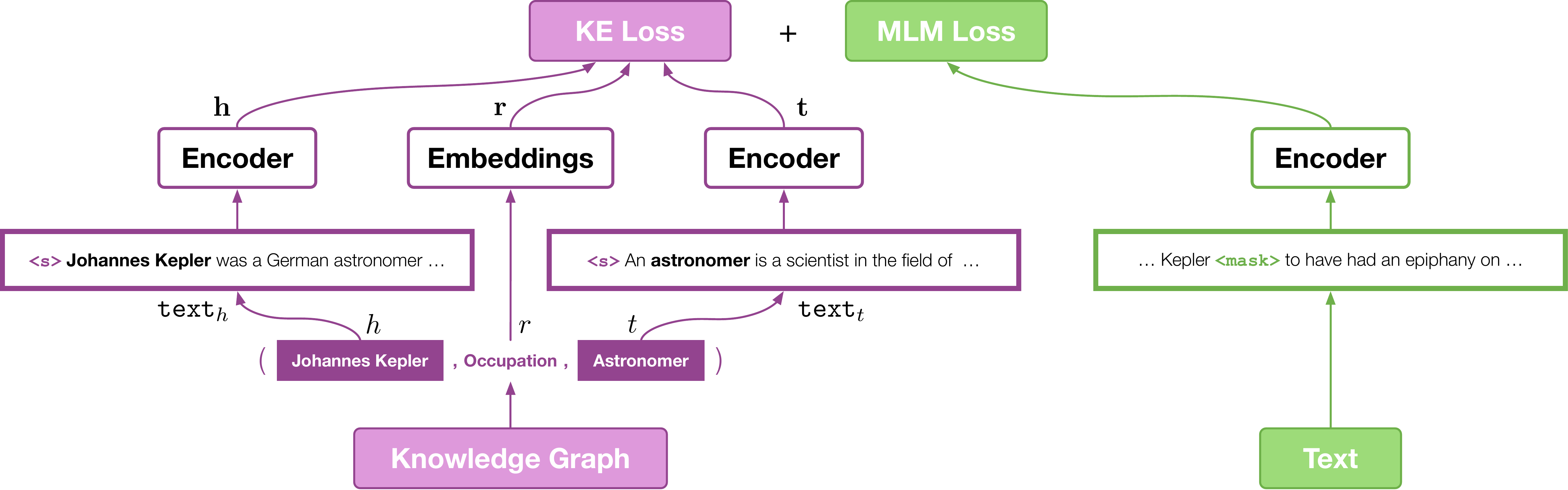}
\caption{The KEPLER framework. We encode entity descriptions as entity embeddings and jointly train the knowledge embedding (KE) and masked language modeling (MLM) objectives on the same PLM. }
\label{fig:kepler}
\end{figure*}



As shown in Figure~\ref{fig:kepler}, KEPLER implicitly incorporates factual knowledge into language representations by jointly training with two objectives. In this section, we detailedly introduce the encoder structure, the KE and MLM objectives, and how we combine the two as a unified model.

\subsection{Encoder}
\label{sec:encoder}
For the text encoder, we use Transformer architecture~\citep{vaswani2017attention} in the same way as~\citet{devlin-etal-2019-bert} and \citet{liu2019roberta}. The encoder takes a sequence of $N$ tokens $(x_1, ..., x_N)$ as inputs, and computes $L$ layers of $d$-dimensional contextualized representations $\mathbf{H}_i\in\mathbb{R}^{N\times d},1\leq i\leq L$. Each layer of the encoder $\texttt{E}_i$ is a combination of a multi-head self-attention network and a multi-layer perceptron, and the encoder gets the representation of each layer by $\mathbf{H}_i=\texttt{E}_i(\mathbf{H}_{i-1})$. Eventually, we get a contextualized representation for each position, which could be further used in downstream tasks. Usually, there is a special token \texttt{<s>} added to the beginning of the text, and the output at \texttt{<s>} is regarded sentence representation. We denote the representation function as $\texttt{E}_{\texttt{<s>}}(\cdot)$.

The encoder requires a tokenizer to convert plain texts into sequences of tokens. Here we use the same tokenization as RoBERTa: the Byte-Pair Encoding (BPE)~\citep{sennrich-etal-2016-neural}.

Unlike previous knowledge-enhanced PLM works~\citep{zhang-etal-2019-ernie,peters-etal-2019-knowledge}, we do not modify the Transformer encoder structure to add external entity linkers or knowledge-integration layers. It means that our model has no additional inference overhead compared to vanilla PLMs, and it makes applying KEPLER in downstream tasks as easy as RoBERTa.

\subsection{Knowledge Embedding}
\label{sec:ke}

To integrate factual knowledge into KEPLER, we adopt the knowledge embedding (KE) objective in our pre-training. 
KE encodes entities and relations in knowledge graphs (KGs) as distributed representations, which benefits lots of downstream tasks, such as link prediction and relation extraction.

We first define KGs: a KG is a graph with entities as its nodes and relations between entities as its edges. We use a triplet $(h,r,t)$ to describe a relational fact, where $h,t$ are the head entity and the tail entity, and $r$ is the relation type within a pre-defined relation set $\mathcal{R}$. 
In conventional KE models, each entity and relation is assigned a $d$-dimensional vector, and a scoring function is defined for training the embeddings and predicting links.

In KEPLER, instead of using stored embeddings, we encode entities into vectors by using their corresponding text. By choosing different textual data and different KE scoring functions, we have multiple variants for the KE objective of KEPLER. In this paper, we explore three simple but effective ways: entity descriptions as embeddings, entity and relation descriptions as embeddings, and entity embeddings conditioned on relations. We leave exploring advanced KE methods as our future work. 

\paragraph{Entity Descriptions as Embeddings} For a relational triplet $(h, r, t)$, we have:

\begin{equation} \label{equ:only_entity}
\begin{aligned}
	\mathbf{h}&=\texttt{E}_{\texttt{<s>}}(\texttt{text}_h),\\
	\mathbf{t}&=\texttt{E}_{\texttt{<s>}}(\texttt{text}_t),\\
	\mathbf{r}&=\mathbf{T}_r,
\end{aligned}
\end{equation}

where $\texttt{text}_h$ and $\texttt{text}_t$ are the descriptions for $h$ and $t$, with a special token $\texttt{<s>}$ at the beginning. 
$\mathbf{T}\in \mathbb{R}^{|\mathcal{R}|\times d}$ is the relation embeddings and $\mathbf{h},\mathbf{t},\mathbf{r}$ are the embeddings for $h,t$ and $r$. 

We use the loss from~\citet{sun2019rotate} as our KE objective, which adopts negative sampling~\citep{mikolov2013distributed} for efficient optimization:

\begin{equation} \label{equ:ke}
\begin{aligned}
    \mathcal{L}_{\texttt{KE}}=-\log{\sigma(\gamma-d_r(\mathbf{h}, \mathbf{t}))}\\
    -\sum_{i=1}^{n}\frac{1}{n}\log{\sigma(d_r(\mathbf{h_i^{\prime}}, \mathbf{t_i^{\prime}})-\gamma)},
\end{aligned}
\end{equation}

where $(h_i^{\prime}, r, t_i^{\prime})$ are negative samples, $\gamma$ is the margin, $\sigma$ is the sigmoid function, and $d_r$ is the scoring function, for which we choose to follow TransE~\citep{bordes2013translating} for its simplicity,

\begin{equation}\label{equ:transe}
    d_r(\mathbf{h}, \mathbf{t})=\|\mathbf{h}+\mathbf{r}-\mathbf{t}\|_p,
\end{equation}

where we take the norm $p$ as $1$. The negative sampling policy is to fix the head entity and randomly sample a tail entity, and vice versa.

\paragraph{Entity and Relation Descriptions as Embeddings} A natural extension for the last method is to encode the relation descriptions as relation embeddings as well. Formally, we have,

\begin{equation}\label{equ:rhat}
	\hat{\mathbf{r}}=\texttt{E}_{\texttt{<s>}}(\texttt{text}_{r}),
\end{equation}

where $\texttt{text}_{r}$ is the description for the relation $r$. Then we use $\hat{\mathbf{r}}$ to replace $\mathbf{r}$ in Equation~\ref{equ:ke} and~\ref{equ:transe}.

\paragraph{Entity Embeddings Conditioned on Relations} In this manner, we use entity embeddings conditioned on $r$ for better KE performances. The intuition is that semantics of an entity may have multiple aspects, and different relations focus on different ones~\citep{lin2015learning}. So we have,

\begin{equation}\label{equ:rel}
	\mathbf{h}_r=\texttt{E}_{\texttt{<s>}}(\texttt{text}_{h,r}),
\end{equation}

where $\texttt{text}_{h,r}$ is the concatenation of the description for the entity $h$ and the description for the relation $r$, with the special token \texttt{<s>} at the beginning and \texttt{</s>} in between. Correspondingly, we use $\mathbf{h}_r$ instead of $\mathbf{h}$ for Equation~\ref{equ:ke} and~\ref{equ:transe}.
\begin{table*}[t]
\centering
\tablefont
\begin{tabular}{lrrrrr}
\toprule
\textbf{Dataset} & \textbf{\#entity} & \textbf{\#relation} & \textbf{\#training} & \textbf{\#validation} & \textbf{\#test} \\ \midrule
FB15K            & $14,951$           & $1,345$              & $483,142$            & $50,000$               & $59,071$         \\
WN18             & $40,943$           & $18$                 & $141,442$            & $5,000$                & $5,000$          \\
FB15K-237        & $14,541$           & $237$                & $272,115$            & $17,535$               & $20,466$         \\
WN18RR           & $40,943$           & $11$                 & $86,835$             & $3,034$                & $3,134$          \\ \midrule
Wikidata5M       & $4,594,485$        & $822$                & $20,614,279$         & $5,163$               & $5,133$         \\ \bottomrule
\end{tabular}
     \caption{Statistics of Wikidata5M (transductive setting) compared with existing KE benchmarks.}
     \label{tab:dataset_statistics}
\end{table*}

\subsection{Masked Language Modeling}
\label{sec:lm}

The masked language modeling (MLM) objective is inherited from BERT and RoBERTa. During pre-training, MLM randomly selects some of the input positions, and the objective is to predict the tokens at these selected positions within a fixed dictionary. 

To be more specific, MLM randomly selects $15\%$ of input positions, among which $80\%$ are masked with the special token \texttt{<mask>}, $10\%$ are replaced by other random tokens, and the rest remain unchanged. For each selected position $j$, the last layer of the contextualized representation $\mathbf{H}_{L,j}$ is used for a $W$-way classification, where $W$ is the size of the dictionary. At last, a cross-entropy loss $\mathcal{L}_{\texttt{MLM}}$ is calculated over these selected positions.

We initialize our model with the pre-trained checkpoint of RoBERTa$_{\small \texttt{BASE}}$\xspace. However, we still keep MLM as one of our objectives to avoid catastrophic forgetting~\citep{mccloskey1989catastrophic} while training towards the KE objective. Actually, as demonstrated in Section~\ref{sec:ablation}, only using the KE objective leads to poor results in NLP tasks.


\subsection{Training Objectives}
\label{sec:objective}

To incorporate factual knowledge and language understanding into one PLM, we design a multi-task loss as shown in Figure \ref{fig:kepler} and Equation \ref{equ:loss},

\begin{equation}
    \mathcal{L}=\mathcal{L}_{\texttt{KE}}+\mathcal{L}_{\texttt{MLM}},
    \label{equ:loss}
\end{equation}

where $\mathcal{L}_{\texttt{KE}}$ and $\mathcal{L}_{\texttt{MLM}}$ are the losses for KE and MLM correspondingly. 
Jointly optimizing the two objectives can implicitly integrate knowledge from external KGs into the text encoder, while preserving the strong abilities of PLMs for syntactic and semantic understanding. Note that those two tasks only share the text encoder, and for each mini-batch, text data sampled for KE and MLM are not (necessarily) the same. This is because seeing a variety of text (instead of just entity descriptions) in MLM can help the model to have better language understanding ability.

\subsection{Variants and Implementations}
\label{sec:pretraining}

We introduce the variants of KEPLER and the pre-training implementations here. The fine-tuning details will be introduced in Section~\ref{sec:exp}.

\subsubsection*{KEPLER Variants}

We implement multiple versions of KEPLER in experiments to explore the effectiveness of our pre-training framework. We use the same denotations in Section~\ref{sec:exp} as below. 

\textbf{KEPLER-Wiki} is the principal model in our experiments, which adopts Wikidata5M (Section~\ref{sec:dataset}) as the KG and the entity-description-as-embedding method (Equation~\ref{equ:only_entity}). 
All other variants, if not specified, use the same settings. 
KEPLER-Wiki achieves the best performances on most tasks. 


\textbf{KEPLER-WordNet} uses the WordNet~\citep{miller1995wordnet} as its KG source. WordNet is an English lexical graph, where nodes are lemmas and synsets, and edges are their relations. Intuitively, incorporating WordNet can bring lexical knowledge and thus benefits NLP tasks. We use the same WordNet 3.0 as in KnowBert~\citep{peters-etal-2019-knowledge}, which is extracted from the \texttt{nltk}\footnote{\url{https://www.nltk.org}} package.

\textbf{KEPLER-W+W} takes both Wikidata5M and WordNet as its KGs. To jointly train with two KG datasets, we modify the objective in Equation~\ref{equ:loss} as
\begin{equation}
	    \mathcal{L}=\mathcal{L}_{\texttt{Wiki}}+\mathcal{L}_{\texttt{WordNet}}+\mathcal{L}_{\texttt{MLM}},
\end{equation}

where $\mathcal{L}_{\texttt{Wiki}}$ and $\mathcal{L}_{\texttt{WordNet}}$ are losses from Wikidata5M and WordNet respectively. 

\textbf{KEPLER-Rel} uses the entity and relation descriptions as embeddings method (Equation~\ref{equ:rhat}). 
As the relation descriptions in Wikidata are short ($11.7$ words on average) and homogeneous, encoding relation descriptions as relation embeddings results in worse performance as shown in Section~\ref{sec:exp}.

\textbf{KEPLER-Cond} uses the entity-embedding-conditioned-on-relation method (Equation~\ref{equ:rel}). 
This model achieves superior results in link prediction tasks, both transductive and inductive (Section~\ref{sec:KEexp}).

\textbf{KEPLER-OnlyDesc} trains the MLM objective directly on the entity descriptions from the KE objective rather than uses the English Wikipedia and BookCorpus as other versions of KEPLER. However, as the entity description data are smaller (2.3~GB vs 13~GB) and homogeneous, it harms the general language understanding ability and thus performs worse~(Section~\ref{sec:nlpexp}).

\textbf{KEPLER-KE} only adopts the KE objective in pre-training, which is an ablated version of KEPLER-Wiki. It is used to show the necessity of the MLM objective for language understanding.

\subsubsection*{Pre-training Implementation}

In practice, we choose RoBERTa~\citep{liu2019roberta} as our base model and implement KEPLER in the fairseq framework~\citep{ott2019fairseq} for pre-training. Due to the computing resource limit, we choose the \texttt{BASE} size ($L=12$, $d=768$) and use the released \texttt{roberta.base} parameters for initialization, which is a common practice to save pre-training time~\citep{zhang-etal-2019-ernie,peters-etal-2019-knowledge}. For the MLM objective, we use the English Wikipedia (2,500M words) and BookCorpus (800M words)~\citep{zhu2015aligning} as our pre-training corpora (except KEPLER-OnlyDesc). We extract text from these two corpora in the same way as~\citet{devlin-etal-2019-bert}. For the KE objective, we encode the first $512$ tokens of entity descriptions from the English Wikipedia as entity embeddings. 

We set the $\gamma$ in Equation~\ref{equ:ke} as $4$ and $9$ for NLP and KE tasks respectively, and we use the models pre-trained with 10 and 30 epochs for NLP and KE. Specially, the $\gamma$ is $1$ for KEPLER-WordNet. The two hyper-parameters are tuned by multiple trials for $\gamma$ in $\{1,2,4,6,9\}$ and the number of epochs in $\{5,10,20,30,40\}$, and we select the model by performances on TACRED (F-1) and inductive link prediction (HITS@10). We use gradient accumulation to achieve a batch size of $12,288$.

\begin{table}
    \tablefont
    \centering
        \begin{tabular}{lrr}
            \toprule
            \textbf{Entity Type}& \textbf{Occurrence}    & \textbf{Percentage}\\
            \midrule
            Human               & $1,517,591$     & $33.0$\%        \\
            Taxon               & $363,882$       & $7.9$\%        \\ 
            Wikimedia list      & $118,823$       & $2.6$\%        \\
            Film                & $114,266$       & $2.5$\%        \\
            Human Settlement    & $110,939$       & $2.4$\%        \\
            \midrule
            Total               & $2,225,501$    & $48.4$\%        \\
            \bottomrule
        \end{tabular}
    \caption{Top-5 entity categories in Wikidata5M.}
    \label{tab:dataset_top_category}
\end{table}

\begin{table}[t]
    \tablefont
    \centering
    \begin{adjustbox}{max width=0.98\linewidth}
\begin{tabular}{lrrr}
\toprule
\textbf{Subset} & \textbf{\#entity} & \textbf{\#relation} & \textbf{\#triplet} \\ 
\midrule
Training        & $4,579,609$         & $822$                 & $20,496,514$         \\
Validation      & $7,374$             & $199$                 & $6,699$              \\
Test            & $7,475$             & $201$                 & $6,894$  \\
\bottomrule
\end{tabular}
    \end{adjustbox}
    \caption{Statistics of Wikidata5M inductive setting.}
    \label{tab:inductive_statistics}
\end{table}

\begin{table*}[t]
    \tablefont
    \centering
    \begin{tabular}{lrrrrr}
        \toprule
        \textbf{Method}     & \textbf{MR}    & \textbf{MRR} & \textbf{HITS@1} & \textbf{HITS@3} & \textbf{HITS@10}   \\
        \midrule
        TransE~\cite{bordes2013translating}  & $109370$    & $25.3$ & $17.0$     & $31.1$     & $39.2$     \\
        DistMult~\cite{yang2015embedding}    & $211030$    & $25.3$ & $20.8$     & $27.8$     & $33.4$     \\
        ComplEx~\cite{trouillon2016complex}  & $244540$    & $28.1$ & $22.8$     & $31.0$     & $37.3$     \\
        SimplE~\cite{kazemi2018simple}       & $115263$    & $29.6$ & $25.2$     & $31.7$     & $37.7$     \\
        RotatE~\cite{sun2019rotate}          & $89459$     & $29.0$ & $23.4$     & $32.2$     & $39.0$     \\
        \bottomrule
    \end{tabular}
    \caption{Performances of different KE models on Wikidata5M (\% except MR).}
    \label{tab:benchmark}
\end{table*}


\section{Wikidata5M}
\label{sec:dataset}

As shown in Section~\ref{sec:kepler}, to train KEPLER, the KG dataset should (1) be large enough, (2) contain high-quality textual descriptions for its entities and relations, 
and (3) have a reasonable inductive setting, 
which most existing KG datasets do not provide. 
Thus, based on Wikidata\footnote{\url{https://www.wikidata.org}} and English Wikipedia\footnote{\url{https://en.wikipedia.org}}, we construct Wikidata5M, a large-scale KG dataset with aligned text descriptions from corresponding Wikipedia pages, and also an inductive test set. 
In the following sections, we first introduce the data collection (Section~\ref{sec:collection}) and the data split (Section~\ref{sec:split}), and then provide the results of representative KE methods on the dataset (Section~\ref{sec:benchmark}).

\subsection{Data Collection}
\label{sec:collection}

We use the July 2019 dump of Wikidata and Wikipedia. For each entity in Wikidata, we align it to its Wikipedia page and extract the first section as its description. Entities with no pages or with descriptions fewer than $5$ words are discarded.

We retrieve all the relational facts in Wikidata. A fact is considered to be valid when both of its entities are not discarded, and its relation has a non-empty page in Wikidata. The final KG contains $4,594,485$ entities, $822$ relations and $20,624,575$ triplets. Statistics of Wikidata5M along with four other widely-used benchmarks are shown in Table~\ref{tab:dataset_statistics}. Top-5 entity categories are listed in Table~\ref{tab:dataset_top_category}. We can see that Wikidata5M is much larger than other KG datasets, covering various domains.

\subsection{Data Split}
\label{sec:split}

For Wikidata5M, we take two different settings: the transductive setting and the inductive setting. 

The \textbf{transductive setting} (shown in Table~\ref{tab:dataset_statistics}) is adopted in most KG datasets, where the entities are shared and the triplet sets are disjoint across training, validation and test. In this case, KE models are expected to learn effective entity embeddings only for the entities in the training set.

In the \textbf{inductive setting} (shown in Table~\ref{tab:inductive_statistics}), the entities and triplets are mutually disjoint across training, validation and test. We randomly sample some connected subgraphs as the validation and test set. In the inductive setting, the KE models should produce embeddings for the unseen entities given side features like descriptions, neighbors, etc. The inductive setting is more challenging and also meaningful in real-world applications, where entities in KGs experience open-ended growth, and the inductive ability is crucial for online KE methods.

Although Wikidata5M contains massive entities and triplets, our validation and test set are not large, which is limited by the standard evaluation method of link prediction (Section~\ref{sec:benchmark}). Each episode of evaluation requires $|\mathcal{E}|\times|\mathcal{T}|\times2$ times of KE score calculation, where $|\mathcal{E}|$ and $|\mathcal{T}|$ are the total number of entities and the number of triplets in test set respectively. As Wikidata5M contains massive entities, the evaluation is very time-consuming, hence we have to limit the test set to thousands of triplets to ensure tractable evaluations. This indicates that large-scale KE urges a more efficient evaluation protocol. We will leave exploring it to future work.

\subsection{Benchmark}
\label{sec:benchmark}

To assess the challenges of Wikidata5M, we benchmark several popular KE models on our dataset in the transductive setting (as they inherently do not support the inductive setting). Because their original implementations do not scale to Wikidata5M, we benchmark these methods with GraphVite~\citep{zhu2019graphvite}, a multi-GPU KE toolkit.

In the transductive setting, for each test triplet $(h,r,t)$, the model ranks all the entities by scoring $(h,r,t^{\prime}), t^{\prime}\in \mathcal{E}$, where $\mathcal{E}$ is the entity set excluding other correct $t$. The evaluation metrics, MRR (mean reciprocal rank), MR (mean rank), and HITS@\{1,3,10\}, are based on the rank of the correct tail entity $t$ among all the entities in $\mathcal{E}$. Then we do the same thing for the head entities. We report the average results over all test triplets and over both head and tail entity predictions.

Table \ref{tab:benchmark} shows the results of popular KE methods on Wikidata5M, which are all significantly lower than on existing KG datasets like FB15K-237, WN18RR, etc. It demonstrates that Wikidata5M is more challenging due to its large scale and high coverage. 
The results advocate for more efforts towards large-scale KE.


\newcommand\ROBERTABASE{RoBERTa$_{\small \texttt{BASE}}$\xspace}
\newcommand\BERTBASE{BERT$_{\small \texttt{BASE}}$\xspace}
\newcommand\BERTLARGE{BERT$_{\small \texttt{LARGE}}$\xspace}
\newcommand\RERNIE{ERNIE$_{\small \texttt{RoBERTa}}$\xspace}
\newcommand\RKNOWBERT{KnowBert$_{\small \texttt{RoBERTa}}$\xspace}
\section{Experiments}
\label{sec:exp}

In this section, we introduce the experiment settings and results of our model on various NLP and KE tasks, along with some analyses on KEPLER.

\subsection{Experimental Setting}
\label{sec:expsetting}

\paragraph{Baselines} 
In our experiments, \textbf{RoBERTa} is an important baseline since KEPLER is based on it (all mentioned models are of \texttt{BASE} size if not specified). 
As we cannot afford the full RoBERTa corpora (126~GB, and we only use 13~GB) in KEPLER pre-training, we implement \textbf{Our RoBERTa} for direct comparisons to KEPLER. It is initialized by \ROBERTABASE and is further trained with the MLM objective on the same corpora as KEPLER.

We also evaluate recent knowledge-enhanced PLMs, including \textbf{ERNIE$_{\small \texttt{BERT}}$}~\citep{zhang-etal-2019-ernie} and \textbf{KnowBert$_{\small \texttt{BERT}}$}~\citep{peters-etal-2019-knowledge}. 
As ERNIE and our principal model KEPLER-Wiki only use Wikidata, we take KnowBert-Wiki in the experiments to ensure fair comparisons with the same knowledge source. Considering KEPLER is based on RoBERTa, we reproduce the two models with RoBERTa too (\textbf{\RERNIE} and \textbf{\RKNOWBERT}). The reproduction of KnowBert is based on its original implementation\footnote{\url{https://github.com/allenai/kb}}. 
On relation classification, we also compare with MTB~\citep{baldini-soares-etal-2019-matching}, which adopts ``matching the blank'' pre-training. Different from other baselines, the original MTB is based on \BERTLARGE (denoted by \textbf{MTB (\BERTLARGE)}). For a fair comparison under the same model size, we reimplement MTB with \BERTBASE (\textbf{MTB}).

\paragraph{Hyper-parameter}
The pre-training settings are in Section~\ref{sec:pretraining}. For fine-tuning on downstream tasks, we set KEPLER hyper-parameters the same as reported in KnowBert on TACRED and OpenEntity. On FewRel, we set the learning rate as $2$e-$5$ and batch size as $20$ and $4$ for the Proto and PAIR frameworks respectively. For GLUE, we follow the hyper-parameters reported in RoBERTa. For baselines, we keep their original hyper-parameters unchanged or use the best trial in KEPLER searching space if no original settings are available.

\subsection{NLP Tasks}
\label{sec:nlpexp}
In this section, we demonstrate the performance of KEPLER and its baselines on various NLP tasks. 

\subsubsection*{Relation Classification}

Relation classification requires models to classify relation types between two given entities from text. We evaluate KEPLER and other baselines on two widely-used benchmarks: TACRED and FewRel.

\begin{table}[t]
    \centering
        \begin{adjustbox}{max width=0.98\linewidth}

        \tablefont
    \begin{tabular}{lccc}
        \toprule
        \textbf{Model} & \textbf{P} & \textbf{R} & \textbf{F-1} \\
        \midrule
        BERT & $67.2$ & $64.8$ & $66.0$ \\
        \BERTLARGE & - & - & $70.1$ \\
        MTB & $69.7$ & $67.9$ & $68.8$ \\
        MTB (\BERTLARGE) & - & - & $71.5$ \\
        ERNIE$_{\small \texttt{BERT}}$ & $70.0$ & $66.1$ & $68.0$ \\
        KnowBert$_{\small \texttt{BERT}}$ & $\bm{73.5}$ & $64.1$ & $68.5$ \\
        RoBERTa & $70.4$ & $71.1$ & $70.7$ \\
        \RERNIE & $\bm{73.5}$ & $68.0$ & $70.7$ \\
        \RKNOWBERT & $71.9$ & $69.9$ & $70.9$ \\ 
        \midrule
        Our RoBERTa & $70.8$ & $69.6$ & $70.2$ \\
        KEPLER-Wiki & $71.5$ & $\bm{72.5}$ & $\bm{72.0}$ \\
        KEPLER-WordNet & $71.4$ & $71.3$ & $71.3$ \\
        KEPLER-W+W & $71.1$ & $72.0$ & $71.5$ \\
        KEPLER-Rel & $71.3$ & $70.9$ & $71.1$ \\
        KEPLER-Cond & $72.1$ & $70.7$ & $71.4$ \\
        KEPLER-OnlyDesc & $72.3$ & $69.1$ & $70.7$ \\
        KEPLER-KE & $63.5$ & $60.5$ & $62.0$ \\
        \bottomrule
    \end{tabular}
    \end{adjustbox}
    \caption{Precision, recall and F-1 on TACRED (\%). KnowBert results are different from the original paper since different task settings are used.}
    \label{tab:tacred}
\end{table}

\textbf{TACRED} \citep{zhang-etal-2017-position} has $42$ relations and $106,264$ sentences. Here we follow the settings of~\citet{baldini-soares-etal-2019-matching}, where we add four special tokens before and after the two entity mentions, and concatenate the representations at the beginnings of the two entities for classification. 
Note that the original KnowBert also takes entity types as inputs, which is different from \citet{zhang-etal-2019-ernie,baldini-soares-etal-2019-matching}. To ensure fair comparisons, we re-evaluate KnowBert with the same setting as other baselines, thus the reported results are different from the original paper. 

\begin{table*}[th]

    \centering
    \tablefont
    \begin{adjustbox}{max width=0.98\linewidth}
    \begin{tabular}{l|cccc|cccc}
        \toprule
        \multicolumn{1}{l|}{\multirow{2}{*}{\textbf{Model}}} & \multicolumn{4}{c|}{\textbf{FewRel 1.0}} & \multicolumn{4}{c}{\textbf{FewRel 2.0}}\\
        & \textbf{5-1} & \textbf{5-5} & \textbf{10-1} & \textbf{10-5} & \textbf{5-1} & \textbf{5-5} & \textbf{10-1} & \textbf{10-5}\\
        \midrule
        MTB (\BERTLARGE)$^\dagger$ & $93.86$ & $97.06$ & $89.20$ & $94.27$ & $-$ & $-$ & $-$ & $-$ \\
        \midrule
        Proto (BERT) & $80.68$ & $89.60$ & $71.48$ & $82.89$ & $40.12$ & $51.50$ & $26.45$ & $36.93$\\
        Proto (MTB) & $81.39$ &	$91.05$&	$71.55$&	$83.47$&	$52.13$&	$76.67$&	$48.28$ &	$69.75$ \\
        Proto (ERNIE$_{\small \texttt{BERT}}$)$^\dagger$ & $\textbf{89.43}$ & $94.66$ & $\textbf{84.23}$ & $90.83$ & $49.40$ & $65.55$ & $34.99$ & $49.68$\\
        Proto (KnowBert$_{\small \texttt{BERT}}$)$^\dagger$ & $86.64$ & $93.22$ & $79.52$ & $88.35$ & $64.40$ & $79.87$ & $51.66$ & $69.71$\\
        Proto (RoBERTa) & $85.78$ & $95.78$ & $77.65$ & $92.26$ & $64.65$ & $82.76$ & $50.80$ & $71.84$\\
        Proto (Our RoBERTa) & $84.42$ & $95.30$ & $76.43$ & $91.74$ & $61.98$ & $83.11$ & $48.56$ & $72.19$ \\
        Proto (\RERNIE)$^\dagger$ & $87.76$ & $95.62$ & $80.14$ & $91.47$ & $54.43$ & $80.48$ & $37.97$ & $66.26$ \\
        Proto (\RKNOWBERT)$^\dagger$ & $82.39$ & $93.62$ & $76.21$ & $88.57$ & $55.68$ & $71.82$ & $41.90$ & $58.55$\\
        Proto (KEPLER-Wiki) & $88.30$ & $\textbf{95.94}$ & $81.10$ & $\textbf{92.67}$ & $\textbf{66.41}$ & $\textbf{84.02}$ & $\textbf{51.85}$ & $\textbf{73.60}$ \\
        \midrule
        PAIR (BERT) & $88.32$ & $93.22$ & $80.63$ & $87.02$& $\textbf{67.41}$ & $78.57$ & $\textbf{54.89}$ & $66.85$ \\
        PAIR (MTB) & $83.01$ & $87.64$ & $73.42$ & $78.47$ & $46.18$ & $70.50$ & $36.92$ & $55.17$\\
        PAIR (ERNIE$_{\small \texttt{BERT}}$)$^\dagger$ & $\textbf{92.53}$ & $94.27$ & $\textbf{87.08}$ & $89.13$ & $56.18$ & $68.97$ & $43.40$ & $54.35$\\
        PAIR (KnowBert$_{\small \texttt{BERT}}$)$^\dagger$ & $88.48$ & $92.75$ & $82.57$ & $86.18$ & $66.05$ & $77.88$ & $50.86$ & $67.19$\\
        PAIR (RoBERTa) & $89.32$ & $93.70$ & $82.49$ & $88.43$ & $66.78$ & $81.84$ & $53.99$ & $70.85$\\
        PAIR (Our RoBERTa) & $89.26$ & $93.71$ & $83.32$ & $89.02$ & $63.22$ & $77.66$ & $49.28$ & $65.97$ \\
        PAIR (\RERNIE)$^\dagger$ & $87.46$ & $94.11$ & $81.68$ & $87.83$ & $59.29$ & $72.91$ & $48.51$ & $60.26$ \\
        PAIR (\RKNOWBERT)$^\dagger$ & $85.05$ & $91.34$ & $76.04$ & $85.25$ & $50.68$ & $66.04$ & $37.10$ & $51.13$\\
        PAIR (KEPLER-Wiki) & $90.31$ & $\textbf{94.28}$ & $85.48$ & $\textbf{90.51}$ & $67.23$ & $\textbf{82.09}$ & $54.32$ & $\textbf{71.01}$ \\ 
        \bottomrule
    \end{tabular}
    \end{adjustbox}

    \caption{Accuracies ($\%$) on the FewRel dataset. $N$-$K$ indicates the $N$-way $K$-shot setting. 
MTB uses the \texttt{LARGE} size and all the other models use the \texttt{BASE} size. $^\dagger$ indicates oracle models which may have seen facts in the FewRel 1.0 test set during pre-training.}
    \label{tab:fewrel}

\end{table*}

From the TACRED results in Table~\ref{tab:tacred}, 
we can observe that: (1) KEPLER-Wiki is the best one among KEPLER variants and significantly outperforms all the baselines, while other versions of KEPLER also achieve good results. It demonstrates the effectiveness of KEPLER on integrating factual knowledge into PLMs. Based on the result, we use KEPLER-Wiki as the principal model in the following experiments. (2) KEPLER-WordNet shows a marginal improvement over Our RoBERTa, while KEPLER-W+W underperforms KEPLER-Wiki. It suggests that pre-training with WordNet only has limited benefits in the KEPLER framework. We will explore how to better combine different KGs in our future work.


\textbf{FewRel}~\citep{han-etal-2018-fewrel} is a few-shot relation classification dataset with $100$ relations and $70,000$ instances, which is constructed with Wikipedia text and Wikidata facts. Furthermore, \citet{gao-etal-2019-fewrel} propose \textbf{FewRel 2.0}, adding a domain adaptation challenge with a new medical-domain test set.

FewRel takes the $N$-way $K$-shot setting. Relations in the training and test sets are disjoint. For every evaluation episode, $N$ relations, $K$ supporting samples for each relation, and several query sentences are sampled from the test set. The models are required to classify queries into one of the $N$ relations only given the sampled $N\times K$ instances.

We use two state-of-the-art few-shot frameworks: \textbf{Proto}~\citep{snell2017prototypical} and \textbf{PAIR}~\citep{gao-etal-2019-fewrel}. We replace the text encoders with our baselines and KEPLER and compare the performance. Since FewRel 1.0 is constructed with Wikidata, we remove all the triplets in its test set from Wikidata5M to avoid information leakage for KEPLER. However, we cannot control the KGs used in our baselines. We mark the models utilizing Wikidata and have information leakage risk with $^\dagger$ in Table~\ref{tab:fewrel}.

\begin{table}[t]
    \tablefont
    \centering
        \begin{adjustbox}{max width=0.98\linewidth}
    \begin{tabular}{lccc}
        \toprule
        \textbf{Model} & \textbf{P} & \textbf{R} & \textbf{F-1} \\
        \midrule
        UFET~\citep{choi-etal-2018-ultra} & $77.4$ & $60.6$ & $68.0$\\
        BERT & $76.4$ & $71.0$ & $73.6$ \\
        ERNIE$_{\small \texttt{BERT}}$ & $78.4$ & $72.9$ & $75.6$ \\
        KnowBert$_{\small \texttt{BERT}}$ & $77.9$ & $71.2$ & $74.4$ \\
        RoBERTa & $77.4$ & $73.6$ & $75.4$ \\
        \RERNIE & $80.3$ & $70.2$ & $74.9$ \\
        \RKNOWBERT & $78.7$ & $72.7$ & $75.6$ \\
        \midrule
        Our RoBERTa & $75.1$ & $73.4$ & $74.3$ \\
        KEPLER-Wiki & $77.8$ & $74.6$ & $\textbf{76.2}$\\
        \bottomrule
    \end{tabular}
    \end{adjustbox}
    \caption{Entity typing results on OpenEntity ($\%$).}
    \label{tab:openentity}
\end{table}

As Table~\ref{tab:fewrel} shows, KEPLER-Wiki achieves the best performance over the \texttt{BASE}-size PLMs in most settings. From the results, we also have some interesting observations: 
(1) RoBERTa consistently outperforms BERT on various NLP tasks~\citep{liu2019roberta}, yet the RoBERTa-based models here are comparable or even worse than BERT-based models in the PAIR framework. Since PAIR uses sentence concatenation, this result may be credited to the next sentence prediction (NSP) objective of BERT. (2) KEPLER brings improvements on FewRel 2.0, while ERNIE and KnowBert even degenerate in most of the settings. It indicates that the paradigms of ERNIE and KnowBert cannot well generalize to new domains which may require much different entity linkers and entity embeddings. On the other hand, KEPLER not only learns better entity representations but also acquires a general ability to extract factual knowledge from the context across different domains. We further verify this in Section~\ref{sec:uos}. 
(3) KnowBert underperforms ERNIE in FewRel while it typically achieves better results on other tasks. This may be because it uses the TuckER~\citep{balazevic-etal-2019-tucker} KE model while ERNIE and KEPLER follow TransE~\citep{bordes2013translating}. We will explore the effects of different KE methods in the future. 

We also have another two observations with regard to ERNIE and MTB: (1) ERNIE performs the best on $1$-shot settings of FewRel 1.0. We believe this is because that the knowledge embedding injection of ERNIE has particular advantages in this case, since it directly brings knowledge about entities. When using 5-shot (supporting text provides more information) and FewRel 2.0 (ERNIE does not have knowledge for biomedical entities), KEPLER outperforms ERNIE. (2) Though MTB (\BERTLARGE) is the state-of-the-art model on FewRel, its \BERTBASE version does not outperform other knowledge-enhanced PLMs, which suggests that using large models contributes much to its gain. We also notice that when combined with PAIR, MTB suffers an obvious performance drop, which may be because its pre-training objective degenerates sentence-pair tasks.

\subsubsection*{Entity Typing}

Entity typing requires to classify given entity mentions into pre-defined types. For this task, we carry out evaluations on OpenEntity~\citep{choi-etal-2018-ultra} following the settings in \citet{zhang-etal-2019-ernie}. OpenEntity has 6 entity types and 2,000 instances for training, validation and test each.

To identify the entity mentions of interest, we add two special tokens before and after the entity spans, and use the representations of the first special tokens for classification. As shown in Table~\ref{tab:openentity}, KEPLER-Wiki achieves state-of-the-art results. Note that the KnowBert results are different from the original paper since we use KnowBert-Wiki here rather than KnowBert-W+W to ensure the same knowledge resource and fair comparisons. KEPLER does not perform linking or entity embedding pre-training like ERNIE and KnowBert, which bring them special advantages in entity span tasks. However, KEPLER still outperforms these baselines, which proves its effectiveness. 


\subsubsection*{GLUE}

The General Language Understanding Evaluation (GLUE)~\citep{wang-etal-2018-glue} collects several natural language understanding tasks 
and is widely used for evaluating PLMs. In general, solving GLUE does not require factual knowledge~\citep{zhang-etal-2019-ernie} and we use it to examine whether KEPLER harms the general language understanding ability.

Table~\ref{tab:glue} shows the GLUE results. We can observe that KEPLER-Wiki is close to Our RoBERTa, suggesting that while incorporating factual knowledge, KEPLER maintains a strong language understanding ability. However, there are significant performance drops of KEPLER-OnlyDesc, which indicates that the small-scale entity description data are not sufficient for training KEPLER with MLM.

For the small datasets STS-B, MRPC and RTE, directly fine-tuning models on them typically result in unstable performance. Hence we fine-tune models on a large-scale dataset (here we use MNLI) first and then further fine-tune them on the small datasets.
The method has been shown to be effective~\citep{wang-etal-2019-tell} and is also used in the original RoBERTa paper~\citep{liu2019roberta}.

\subsection{KE Tasks}
\label{sec:KEexp}
We show how KEPLER works as a KE model, and evaluate it on Wikidata5M in both the transductive link prediction setting and the inductive setting.

\subsubsection*{Experimental Settings}
In link prediction, the entity and relation embeddings of KEPLER are obtained as described in Section~\ref{sec:ke} and \ref{sec:pretraining}. The evaluation method is described in Section~\ref{sec:benchmark}.  
We also add RoBERTa and Our RoBERTa as baselines. They adopt Equation~\ref{equ:only_entity} and \ref{equ:rhat} to acquire entity and relation embeddings, and use Equation~\ref{equ:transe} as their scoring function.

In the transductive setting, we compare our models with \textbf{TransE}~\citep{bordes2013translating}. We set its dimension as $512$, negative sampling size as $64$, batch size as $2048$ and learning rate as $0.001$ after hyper-parameter searching.  The negative sampling size is crucial for the performance on KE tasks, but limited by the model complexity, KEPLER can only take a negative size of $1$. For a direct comparison to intuitively show the benefits of pre-training, we set a baseline \textbf{TransE$^\dagger$}, which also uses $1$ as the negative sampling size and keeps the other hyper-parameters unchanged.
    
Since conventional KE methods like TransE inherently cannot provide embeddings for unseen entities, we take \textbf{DKRL}~\citep{Xie:2016:RLK:3016100.3016273} as our baseline in the KE experiments, which utilizes convolutional neural networks to encode entity descriptions as embeddings. We set its dimension as $768$, negative sampling size as $64$, batch size as $1024$ and learning rate as $0.0005$.

\begin{table}[t]
    \centering
    \tablefont
    \begin{adjustbox}{max width=0.98\linewidth}
    \begin{tabular}{lcccc}
        \toprule
        \multirow{2}{*}{\textbf{Model}} & \textbf{MNLI (m/mm)} & \textbf{QQP} &  \textbf{QNLI}  & \textbf{SST-2} \\
              & {\scriptsize \textbf{392K}} & {\scriptsize \textbf{363K}} & {\scriptsize \textbf{104K}} & {\scriptsize \textbf{67K}}\\
        \midrule
        RoBERTa & 87.5/87.2 &  91.9 & 92.7 & 94.8\\
        Our RoBERTa & 87.1/86.8 & 90.9 & 92.5 & 94.7 \\
        KEPLER-Wiki & 87.2/86.5  & 91.7 & 92.4 & 94.5\\
        KEPLER-OnlyDesc & 85.9/85.6  & 90.8 & 92.4 & 94.4\\
        \bottomrule
        \toprule
        \multirow{2}{*}{\textbf{Model}} &  \textbf{CoLA} & \textbf{STS-B} & \textbf{MRPC} & \textbf{RTE} \\
              & {\scriptsize \textbf{8.5K}} & {\scriptsize \textbf{5.7K}} & {\scriptsize \textbf{3.5K}} & {\scriptsize \textbf{2.5K}}  \\
        \midrule
        RoBERTa &   63.6 & 91.2   & 90.2 & 80.9\\
        Our RoBERTa &  63.4 & 91.1  & 88.4 & 82.3\\
        KEPLER-Wiki &  63.6 & 91.2 & 89.3 & 85.2  \\
        KEPLER-OnlyDesc & 55.8 & 90.2 & 88.5 & 78.3\\
        \bottomrule
    \end{tabular}
    \end{adjustbox}
    
    \caption{GLUE results on the \texttt{dev} set (\%). All the results are medians over $5$ runs. We report F-1 scores for QQP and MRPC, Spearman correlations for STS-B, and accuracy scores for the other tasks. The ``m/mm'' stands for matched/mismatched evaluation sets for MNLI~\citep{N18-1101}.} 
    \label{tab:glue}
\end{table}

\begin{table*}[t]
    \tablefont
	\begin{subtable}[h]{0.98\textwidth}
	\centering
	\begin{adjustbox}{max width=0.98\linewidth}
	\begin{tabular}{lrrrrr}
		\toprule
		\textbf{Model} & \textbf{MR} & \textbf{MRR} & \textbf{HITS@1} & \textbf{HITS@3} & \textbf{HITS@10}\\
		\midrule
		TransE \cite{bordes2013translating} & $109370$    & $\textbf{25.3}$ & $17.0$     & $\textbf{31.1}$     & $\textbf{39.2}$     \\
		TransE$^\dagger$ & $406957$ & $6.0$ & $1.8$ & $8.0$ & $13.6$ \\
		DKRL \citep{Xie:2016:RLK:3016100.3016273} & $31566$ & $16.0$ & $12.0$ & $18.1$ & $22.9$  \\
		RoBERTa & $1381597$ & $0.1$ & $0.0$ & $0.1$ & $0.3$ \\
		Our RoBERTa & $1756130$ & $0.1$ & $0.0$ & $0.1$ & $0.2$ \\
		KEPLER-KE & $76735$ & $8.2$ & $4.9$ & $8.9$ & $15.1$ \\
		KEPLER-Rel & $15820$ & $6.6$ & $3.7$ & $7.0$ & $11.7$ \\
		KEPLER-Wiki & $\textbf{14454}$ & $15.4$ & $10.5$ & $17.4$ & $24.4$ \\
		KEPLER-Cond & $20267$ & $21.0$ & $\textbf{17.3}$ & $22.4$ & $27.7$  \\
		\bottomrule
	\end{tabular}
	\end{adjustbox}
	\caption{Transductive results on Wikidata5M (\% except MR). TransE$^\dagger$ denotes a TransE modeled trained with the same negative sampling size ($1$) as KEPLER.}
	\label{tab:trans}
	\end{subtable}

	\medskip
	
		\begin{subtable}[h]{0.98\textwidth}
	\centering
	\begin{tabular}{lrrrrr}
		\toprule
		\textbf{Model} & \textbf{MR} & \textbf{MRR} & \textbf{HITS@1} & \textbf{HITS@3} & \textbf{HITS@10}\\
		\midrule
		DKRL \citep{Xie:2016:RLK:3016100.3016273} & 78 & 23.1 & 5.9 & 32.0 & 54.6\\
		RoBERTa & $723$ & $7.4$ & $0.7$ & $1.0$ & $19.6$ \\
		Our RoBERTa & $1070$ & $5.8$ & $1.9$ & $6.3$ & $13.0$ \\
		KEPLER-KE & $138$ & $17.8$ & $5.7$ & $22.9$ & $40.7$ \\
		KEPLER-Rel & $35$ & $33.4$ & $15.9$ & $43.5$ & $66.1$ \\
		KEPLER-Wiki & $32$ & $35.1$ & $15.4$ & $46.9$ & $71.9$ \\
		KEPLER-Cond & $\textbf{28}$ & $\textbf{40.2}$ & $\textbf{22.2}$ & $\textbf{51.4}$ & $\textbf{73.0}$ \\
		\bottomrule
	\end{tabular}
	\caption{Inductive results on Wikidata5M (\% except MR).}
	\label{tab:induct}
	\end{subtable}
	\caption{Link prediction results on Wikidata5M transductive and inductive settings. }
	\label{tab:keplerkg}
\end{table*}

\subsubsection*{Transductive Setting}

Table~\ref{tab:trans} shows the results of the transductive setting. We observe that: 

(1) KEPLER underperforms TransE. It is reasonable since KEPLER is limited by its large model size, and thus cannot use a large negative sampling size ($1$ for KEPLER, while typical KE methods use $64$ or more) and more training epochs ($30$ vs $1000$ for TransE), which are crucial for KE~\citep{zhu2019graphvite}. On the other hand, KEPLER and its variants perform much better than TransE$^\dagger$ (with a negative sampling size of $1$), showing that using the same negative sampling size, KEPLER can benefit from pre-trained language representations and textual entity descriptions so that outperform TransE. In the future, we will explore reducing the model size of KEPLER to take advantage of both large negative sampling size and pre-training.

(2) The vanilla RoBERTa perform poorly in KE while KEPLER achieves favorable performances, which demonstrates the effectiveness of our multi-task pre-training to infuse factual knowledge. 

(3) Among the KEPLER variants, KEPLER-Cond has superior results, which substantiates the intuition in Section~\ref{sec:ke}. KEPLER-Rel performs worst, which we believe is due to the short and homogeneous relation descriptions of Wikidata. KEPLER-KE significantly underperforms KEPLER-Wiki, which suggests that the MLM objective is necessary as well for the KE tasks to build effective language representation.

(4) We also notice that DKRL performs well on the transductive setting and the result is close to KEPLER. We believe this is because DKRL takes a much smaller encoder (CNN) and thus is easier to train. In the more difficult inductive setting, the gap between DKRL and KEPLER is larger, which better shows the language understanding ability of KEPLER to utilize textual entity descriptions.

\subsubsection*{Inductive Setting}

Table \ref{tab:induct} shows the Wikidata5M inductive results. KEPLER outperforms DKRL and RoBERTa by a large margin, demonstrating the effectiveness of our joint training method. But KEPLER results are still far from ideal performances required by practical applications (constructing KG from scratch, etc.), which urges further efforts on inductive KE. Comparisons among KEPLER variants are consistent with in the transductive setting. 

In addition, we clarify why results in the inductive setting are much higher than the transductive setting, while the inductive setting is more difficult: As shown in Table~\ref{tab:dataset_statistics} and~\ref{tab:inductive_statistics}, the entities involved in the inductive evaluation is much less than the transductive setting ($7,475$ vs. $4,594,485$). Considering the KE evaluation metrics are based on entity ranking, it is reasonable to see higher values in the inductive setting. The performance in different settings should not be directly compared.


\begin{table}[t]
    \tablefont
	\centering
	\begin{adjustbox}{max width=0.98\linewidth}
	\begin{tabular}{lccc}
	\toprule
		\textbf{Model} & \textbf{P} & \textbf{R} & \textbf{F-1}\\
	\midrule
		Our RoBERTa & $70.8$ & $69.6$ & $70.2$\\
		KEPLER-KE & $63.5$ & $60.5$ & $62.0$ \\
		KEPLER-Wiki & $71.5$ & $72.5$ & $72.0$\\
	\bottomrule
	\end{tabular}
	\end{adjustbox}
	\caption{Ablation study results on TACRED (\%).}
	\label{tab:abl}
\end{table}
\section{Analysis}
In this section, we analyze the effectiveness and efficiency of KEPLER with experiments. All the hyper-parameters are the same as reported in Section~\ref{sec:expsetting}, including models in the ablation study.

\subsection{Ablation Study}
\label{sec:ablation}
As shown in Equation~\ref{equ:loss}, KEPLER takes a multi-task loss. To demonstrate the effectiveness of the joint objective, we compare full KEPLER with models trained with only the MLM loss (\textbf{Our RoBERTa}) and only the KE loss (\textbf{KEPLER-KE}) on TACRED. As demonstrated in Table~\ref{tab:abl}, compared to KEPLER-Wiki, both ablation models suffer significant drops. It suggests that the performance gain of KEPLER is credited to the joint training towards both objectives. 

\subsection{Knowledge Probing Experiment}

Section~\ref{sec:nlpexp} shows that KEPLER can achieve significant improvements on NLP tasks requiring factual knowledge. To further verify whether KEPLER can better integrate factual knowledge into PLMs and help to recall them, we conduct experiments on LAMA~\citep{petroni2019language}, a widely-used knowledge probe. LAMA examines PLMs' abilities on recalling relational facts by cloze-style questions. For instance, given a natural language template ``Paris is the capital of \texttt{<mask>}'', PLMs are required to predict the masked token without fine-tuning. 
LAMA reports the micro-averaged precision at one (P@1) scores. However, \citet{Poerner2019EBERT} present that LAMA contains some easy questions which can be answered with superficial clues like entity names. Hence we also evaluate the models on LAMA-UHN~\citep{Poerner2019EBERT}, which filters out the questionable templates from the Google-RE and T-REx corpora of LAMA.

\begin{table*}[t]
\centering
\begin{adjustbox}{max width=0.98\linewidth}
\begin{tabular}{l|cccc|cc}
\toprule
\multirow{2}{*}{\textbf{Model}} & \multicolumn{4}{c|}{\textbf{LAMA}}                                         & \multicolumn{2}{c}{\textbf{LAMA-UHN}} \\ \cmidrule{2-7} 
                                & \textbf{Google-RE} & \textbf{T-REx} & \textbf{ConceptNet} & \textbf{SQuAD} & \textbf{Google-RE}  & \textbf{T-REx}  \\ \midrule
BERT                       & $9.8$              & $31.1$         & $15.6$              & $14.1$         &  $4.7$         & $21.8$              \\
RoBERTa                         & $5.3$              & $24.7$         & $19.5$              & $9.1$          & $2.2$               & $17.0$          \\ \midrule
Our RoBERTa                     & $7.0$              & $23.2$         & $\bm{19.0}$              & $8.0$          & $2.8$               & $15.7$          \\
KEPLER-Wiki                     & $\textbf{7.3}$              & $\textbf{24.6}$         & $18.7$              & $\textbf{14.3}$         & $3.3$               & $16.5$          \\
KEPLER-W+W                      & $\textbf{7.3}$              & $24.4$         & $17.6$              & $10.8$         & $\textbf{4.1}$               & $\textbf{17.1}$          \\ \bottomrule
\end{tabular}
\end{adjustbox}
	\caption{P@1 results on knowledge probing benchmark LAMA and LAMA-UHN.}
	\label{tab:lama}
\end{table*}

The evaluation results are shown in Table~\ref{tab:lama}, from which we have the following observations: (1)~KEPLER consistently outperforms the vanilla PLM baseline Our RoBERTa in almost all the settings except ConceptNet, which focuses on commonsense knowledge rather than factual knowledge. It indicates that KEPLER can indeed better integrate factual knowledge. (2) Although KEPLER-W+W cannot outperform KEPLER-Wiki on NLP tasks (Section~\ref{sec:nlpexp}), it shows significant improvements in LAMA-UHN, which suggests that we should explore which kind of knowledge is needed on different scenarios in the future. (3) All the RoBERTa-based models perform worse than vanilla \BERTBASE by a large margin, which is consistent with the results of~\citet{Wang2020KAdapter}. This may be due to different vocabularies used in BERT and RoBERTa, which presents the vulnerability of LAMA-style probing again~\citep{kassner-schutze-2020-negated}. We will leave developing a better knowledge probing framework as our future work.

\subsection{Running Time Comparison}

\begin{table}[t]
\centering
\begin{adjustbox}{max width=0.98\linewidth}
\begin{tabular}{lccc}
\toprule
\textbf{Model}    & \textbf{Entity} & \textbf{Fine-} & \textbf{Inference} \\
\textbf{}    & \textbf{Linking} & \textbf{tuning} & \textbf{} \\ \midrule
\RERNIE    & 780s           & 730s        & 194s      \\
\RKNOWBERT & 190s           & 677s        & 235s      \\
KEPLER   & \textbf{0s}    & \textbf{508s} & \textbf{152s}    \\ \bottomrule
\end{tabular}
\end{adjustbox}
\caption{Three parts of running time for one epoch of TACRED training set.} 
\label{tab:time}
\end{table}

Compared to vanilla PLMs, KEPLER does not introduce any additional parameters or computations during fine-tuning and inference, which is efficient for practice use. 
We compare the running time of KEPLER and other knowledge-enhanced PLMs (ERNIE and KnowBert) in Table~\ref{tab:time}. The time is evaluated on TACRED training set for one epoch with one NVIDIA Tesla V100 (32~GB), and all models use $32$ batch size and $128$ sequence length. The ``entity linking'' time of KnowBert is for entity candidate generation. We can observe that KEPLER requires much less running time since it does not need entity linking or entity embedding fusion, which will benefit time-sensitive applications. 

\subsection{Correlation with Entity Frequency}
\label{sec:entFreq}

\begin{table*}[t]
\centering
    \begin{adjustbox}{max width=0.98\linewidth}
\begin{tabular}{l|ccccc}
\toprule
\textbf{Entity Frequency} & \textbf{0\%-20\%} & \textbf{20\%-40\%} & \textbf{40\%-60\%} & \textbf{60\%-80\%} & \textbf{80\%-100\%} \\ \midrule
KEPLER-Wiki      & $64.7$     & $64.4$      & $64.8$     & $64.7$      & $68.8$       \\
Our RoBERTa          & $64.1$     & $64.3$      & $64.5$      & $64.3$      & $68.5$       \\ \midrule
Improvement      & $+0.6$     & $+0.1$      & $+0.3$      & $+0.4$      & $+0.3$       \\ \bottomrule
\end{tabular}
\end{adjustbox}
\caption{F-1 scores on TACRED (\%) under different settings by entity frequencies. We sort the entity mentions in TACRED by their corresponding entity frequencies in Wikipedia. The ``0\%-20\%'' setting indicates only keeping the least frequent 20\% entity mentions and masking all the other entity mentions (for both training and validation), and so on. The results are averaged over 5 runs.}
\label{tab:freq}
\end{table*}

\begin{figure}[t]
\centering
\includegraphics[width = 0.48\textwidth]{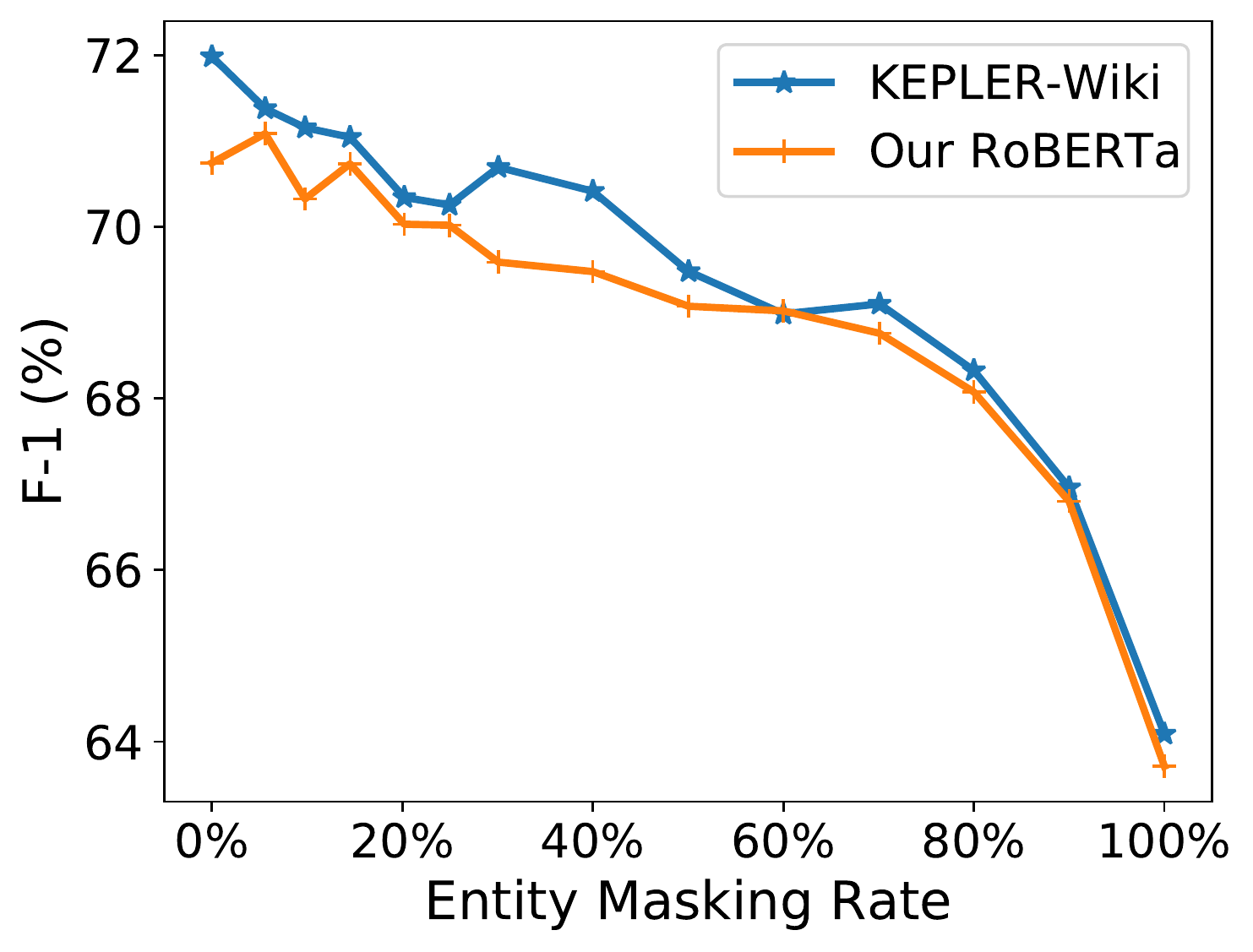}
\caption{TACRED performance (F-1) of KEPLER and RoBERTa change with the rate of entity mentions being masked.}
\label{fig:entFreq}
\end{figure}

To better understand how KEPLER helps the entity-centric tasks, we provide analyses on the correlations between KEPLER performance and entity frequency in this section. The motivation is to verify a natural hypothesis that KEPLER improvements mainly come from better representing the entity mentions in text, especially the rare entities, which do not show up frequently in the pre-training corpora and thus cannot be well learned by the language modeling objectives. 

We perform entity linking for the TACRED dataset with BLINK~\cite{wu2019zero} to link the entity mentions in text to their corresponding Wikipedia identifiers. Then we count the occurrences of the entities in Wikipedia with the hyperlinks in rich text, denoting the entity frequencies. We conduct two experiments to analyze the correlations between KEPLER performance and entity frequency: (1) In Table~\ref{tab:freq}, we divide the entity mentions into five parts by their frequencies, and compare the TACRED performances while only keeping entities in one part and masking the other. (2) In Figure~\ref{fig:entFreq}, we sequentially mask the entity mentions in the ascending order of entity frequencies and see the F-1 changes.

From the results, we can observe that: 

(1) Figure~\ref{fig:entFreq} shows that when the entity masking rate is low, the improvements of KEPLER over RoBERTa are generally much higher than when the entity masking rate is high. It indicates that the improvements of KEPLER do mainly come from better modeling entities in context. However, even when all the entity mentions are masked, KEPLER still outperforms RoBERTa. We claim this is because the KE objective can also help to learn to understand fact-related text since it requires the model to recall facts from textual descriptions. This claim is further substantiated in Section~\ref{sec:uos}. 

(2) From Table~\ref{tab:freq}, we can observe that the improvement in the ``0\%-20\%'' setting is marginally higher than the other settings, which demonstrates that KEPLER does have special advantages on modeling rare entities compared to vanilla PLMs. But the improvements in the frequent settings are also significant and we cannot say that the overall improvements of KEPLER are mostly from the rare entities. In general, the results in Table~\ref{tab:freq} show that KEPLER can better model all the entities, no matter rare or frequent.

\begin{table}[h]
    \tablefont
	\centering
\begin{adjustbox}{max width=0.98\linewidth}
    \begin{tabular}{lcc}
        \toprule
        \textbf{Model} & \textbf{ME} & \textbf{OE}\\
        \midrule
        Our RoBERTa & $54.0$ & $46.8$\\
        KEPLER-KE & $40.2$ & $47.0$ \\
        KEPLER-Wiki & $54.8$ & $48.9$\\
        \bottomrule
    \end{tabular}
\end{adjustbox}
	\caption{Masked-entity (ME) and only-entity (OE) F-1 scores on TACRED (\%).}
	\label{tab:casestudy}
\end{table}

\subsection{Understanding Text or Storing Knowledge}

\label{sec:uos}

We argue that by jointly training the KE and the MLM objectives, KEPLER (1) can better understand fact-related text and better extract knowledge from text, and also (2) can remember factual knowledge. To investigate the two abilities of KEPLER in a quantitative aspect, we carry out an experiment on TACRED, in which the head and tail entity mentions are masked (masked-entity, ME) or only head and tail entity mentions are shown (only-entity, OE). The ME setting shows to what extent the models can extract facts only from the textual context without the clues in entity names. The OE setting demonstrates to what extent the models can store and predict factual knowledge, as only the entity names are given to the models.

As shown in Table~\ref{tab:casestudy}, KEPLER-Wiki shows significant improvements over Our RoBERTa in both settings, which suggests that KEPLER has indeed possessed superior abilities on both extracting and storing knowledge compared to vanilla PLMs without knowledge infusion. And the KEPLER-KE model performs poorly on the ME setting but achieves marginal improvements on the OE setting. It indicates that without the help of the MLM objective, KEPLER only learns the entity description embeddings and degenerates in general language understanding, while it can still remember knowledge into entity names to some extent.

\section{Related Work}


\paragraph{Pre-training in NLP} There has been a long history of pre-training in NLP. Early works focus on distributed word representations~\citep{collobert2008unified, mikolov2013distributed, pennington-etal-2014-glove}, many of which are often adopted in current models as word embeddings. These pre-trained embeddings can capture the semantics of words from large-scale corpora and thus benefit NLP applications.~\citet{peters-etal-2018-deep} push this trend a step forward by using a bidirectional LSTM to form contextualized word embeddings (ELMo) for richer semantic meanings under different circumstances.

Apart from word embeddings, there is another trend exploring pre-trained language models.~\citet{dai2015semi} propose to train an auto-encoder on unlabeled textual data and then fine-tune it on downstream tasks. \citet{howard-ruder-2018-universal} propose a universal language model (ULMFiT). With the powerful Transformer architecture~\citep{vaswani2017attention},~\citet{radford2018improving} demonstrate an effective pre-trained generative model (GPT). Later,~\citet{devlin-etal-2019-bert} release a pre-trained deep Bidirectional Encoder Representation from Transformers (BERT), achieving state-of-the-art performance on a wide range of NLP benchmarks.

After BERT, similar PLMs spring up recently. \citet{yang2019xlnet} propose a permutation language model (XLNet). Later,~\citet{liu2019roberta} show that more data and more parameter tuning can benefit PLMs, and release a new state-of-the-art model (RoBERTa). Other works explore how to add more tasks~\citep{liu-etal-2019-multi} and more parameters~\citep{raffel2019exploring, lan2019albert} to PLMs.

\paragraph{Knowledge-Enhanced PLMs} Recently, many works have investigated how to incorporate knowledge into PLMs. MTB~\citep{baldini-soares-etal-2019-matching} takes a straightforward ``matching the blank'' pre-training objective to help the relation classification task. ERNIE~\citep{zhang-etal-2019-ernie} identifies entity mentions in text and links pre-processed knowledge embeddings to the corresponding positions, which shows improvements on several NLP benchmarks. With a similar idea as ERNIE, KnowBert~\citep{peters-etal-2019-knowledge} incorporates an integrated entity linker in their model and adopts end-to-end training. Besides, \citet{logan-etal-2019-baracks} and \citet{hayashi2019latent} utilize relations between entities inside one sentence to train better generation models. \citet{xiong2019pretrained} adopt entity replacement knowledge learning for improving entity-related tasks. 

Some contemporaneous or following works try to inject factual knowledge into PLMs in different ways. E-BERT~\citep{Poerner2019EBERT} aligns entity embeddings with word embeddings and then directly adds the aligned embeddings into BERT to avoid additional pre-training. K-Adapter~\citep{Wang2020KAdapter} injects knowledge with additional neural adapters to support continuous learning. 



\paragraph{Knowledge Embedding} KE methods have been extensively studied. Conventional KE models define different scoring functions for relational triplets. For example, TransE~\citep{bordes2013translating} treats tail entities as translations of head entities and uses $L_1$-norm or $L_2$-norm to score triplets, while DistMult~\citep{yang2015embedding} uses matrix multiplications and ComplEx~\citep{trouillon2016complex} adopts complex operations based on it. RotatE~\citep{sun2019rotate} combines the advantages of both of them. 

\paragraph{Inductive Embedding} Above KE methods learn entity embeddings only from KG and are inherently transductive, while some works~\citep{wang2014knowledge,Xie:2016:RLK:3016100.3016273,yamada2016joint,cao2017bridge,shi2018open,cao2018joint} incorporate textual metadata such as entity names or descriptions to enhance the KE methods and hence can do inductive KE to some extent. Besides KG, it is also common for general inductive graph embedding methods~\citep{hamilton2017inductive,bojchevski2017deep} to utilize additional node features like text attributes, degrees, etc. KEPLER follows this line of studies and takes full advantage of textual information with an effective PLM.  

\citet{hamaguchi2017knowledge} and \citet{Wang2019LogicAB} perform inductive KE by aggregating the trained embeddings of the known neighboring nodes with graph neural networks, and thus do not need additional features. But these methods require the unseen nodes to be surrounded by known nodes and cannot embed new (sub)graphs. We leave how to develop KEPLER to do fully inductive KE without additional features as future work.

\section{Conclusion and Future Work}

In this paper, we propose KEPLER, a simple but effective unified model for knowledge embedding and pre-trained language representation. We train KEPLER with both the KE and MLM objectives to align the factual knowledge and language representation into the same semantic space, and experimental results on extensive tasks demonstrate its effectiveness on both NLP and KE applications. Besides, we propose Wikidata5M, a large-scale KG dataset to facilitate future research.

In the future, we will (1) explore advanced ways for more smoothly unifying the two semantic space, including different KE forms and different training objectives, and (2) investigate better knowledge probing methods for PLMs to shed light on knowledge-integrating mechanisms.

\section*{Acknowledgement}
This work is supported by the National Key Research and Development Program of China (No. 2018YFB1004503), the National Natural Science Foundation of China (NSFC No. U1736204, 61533018, 61772302, 61732008), grants from Institute for Guo Qiang, Tsinghua University (2019GQB0003) and Beijing Academy of Artificial Intelligence (BAAI2019ZD0502). Prof. Jian Tang is supported by the Natural Sciences and Engineering Research Council (NSERC) Discovery Grant and the Canada CIFAR AI Chair Program. Xiaozhi Wang and Tianyu Gao are supported by Tsinghua University Initiative Scientific Research Program.  We also thank our action editor, Prof. Doug Downey, and the anonymous reviewers for their consistent help and insightful suggestions.


\bibliographystyle{acl_natbib}
\interlinepenalty=10000
\bibliography{tacl2018}

\begin{thebibliography}{62}
\expandafter\ifx\csname natexlab\endcsname\relax\def\natexlab#1{#1}\fi

\bibitem[{Balazevic et~al.(2019)Balazevic, Allen, and
  Hospedales}]{balazevic-etal-2019-tucker}
Ivana Balazevic, Carl Allen, and Timothy Hospedales. 2019.
\newblock \href {https://doi.org/10.18653/v1/D19-1522} {{T}uck{ER}: {Tensor
  Factorization for Knowledge Graph Completion}}.
\newblock In \emph{Proceedings of EMNLP-IJCNLP}, pages 5185--5194.

\bibitem[{Baldini~Soares et~al.(2019)Baldini~Soares, FitzGerald, Ling, and
  Kwiatkowski}]{baldini-soares-etal-2019-matching}
Livio Baldini~Soares, Nicholas FitzGerald, Jeffrey Ling, and Tom Kwiatkowski.
  2019.
\newblock \href {https://doi.org/10.18653/v1/P19-1279} {{Matching the Blanks:
  Distributional Similarity for Relation Learning}}.
\newblock In \emph{Proceedings of ACL}, pages 2895--2905.

\bibitem[{Bojchevski and G{\"u}nnemann(2018)}]{bojchevski2017deep}
Aleksandar Bojchevski and Stephan G{\"u}nnemann. 2018.
\newblock \href {https://openreview.net/forum?id=r1ZdKJ-0W} {{Deep Gaussian
  Embedding of Graphs: Unsupervised Inductive Learning via Ranking}}.
\newblock In \emph{Proceedings of ICLR}.

\bibitem[{Bordes et~al.(2013)Bordes, Usunier, Garcia-Duran, Weston, and
  Yakhnenko}]{bordes2013translating}
Antoine Bordes, Nicolas Usunier, Alberto Garcia-Duran, Jason Weston, and Oksana
  Yakhnenko. 2013.
\newblock \href
  {https://proceedings.neurips.cc/paper/2013/file/1cecc7a77928ca8133fa24680a88d2f9-Paper.pdf}
  {{Translating Embeddings for Modeling Multi-relational Data}}.
\newblock In \emph{Advances in Neural Information Processing Systems (NIPS)},
  pages 2787--2795.

\bibitem[{Cao et~al.(2018)Cao, Hou, Li, Liu, Li, Chen, and Dong}]{cao2018joint}
Yixin Cao, Lei Hou, Juanzi Li, Zhiyuan Liu, Chengjiang Li, Xu~Chen, and Tiansi
  Dong. 2018.
\newblock \href {https://doi.org/10.18653/v1/D18-1021} {{Joint Representation
  Learning of Cross-lingual Words and Entities via Attentive Distant
  Supervision}}.
\newblock In \emph{Proceedings of EMNLP}, pages 227--237.

\bibitem[{Cao et~al.(2017)Cao, Huang, Ji, Chen, and Li}]{cao2017bridge}
Yixin Cao, Lifu Huang, Heng Ji, Xu~Chen, and Juanzi Li. 2017.
\newblock \href {https://doi.org/10.18653/v1/P17-1149} {{Bridge Text and
  Knowledge by Learning Multi-Prototype Entity Mention Embedding}}.
\newblock In \emph{Proceedings of ACL}, pages 1623--1633.

\bibitem[{Choi et~al.(2018)Choi, Levy, Choi, and
  Zettlemoyer}]{choi-etal-2018-ultra}
Eunsol Choi, Omer Levy, Yejin Choi, and Luke Zettlemoyer. 2018.
\newblock \href {https://doi.org/10.18653/v1/P18-1009} {{Ultra-Fine Entity
  Typing}}.
\newblock In \emph{Proceedings of ACL}, pages 87--96.

\bibitem[{Collobert and Weston(2008)}]{collobert2008unified}
Ronan Collobert and Jason Weston. 2008.
\newblock \href {https://doi.org/10.1145/1390156.1390177} {A unified
  architecture for natural language processing: Deep neural networks with
  multitask learning}.
\newblock In \emph{Proceedings of ICML}, pages 160--167.

\bibitem[{Dai and Le(2015)}]{dai2015semi}
Andrew~M Dai and Quoc~V Le. 2015.
\newblock \href
  {https://papers.nips.cc/paper/5949-semi-supervised-sequence-learning.pdf}
  {Semi-supervised sequence learning}.
\newblock In \emph{Advances in Neural Information Processing Systems (NIPS)},
  pages 3079--3087.

\bibitem[{Devlin et~al.(2019)Devlin, Chang, Lee, and
  Toutanova}]{devlin-etal-2019-bert}
Jacob Devlin, Ming-Wei Chang, Kenton Lee, and Kristina Toutanova. 2019.
\newblock \href {https://doi.org/10.18653/v1/N19-1423} {{BERT}: {Pre-training
  of Deep Bidirectional Transformers for Language Understanding}}.
\newblock In \emph{Proceedings of NAACL-HLT}, pages 4171--4186.

\bibitem[{Gao et~al.(2019)Gao, Han, Zhu, Liu, Li, Sun, and
  Zhou}]{gao-etal-2019-fewrel}
Tianyu Gao, Xu~Han, Hao Zhu, Zhiyuan Liu, Peng Li, Maosong Sun, and Jie Zhou.
  2019.
\newblock \href {https://www.aclweb.org/anthology/D19-1649} {{F}ew{R}el 2.0:
  {Towards More Challenging Few-Shot Relation Classification}}.
\newblock In \emph{Proceedings of EMNLP-IJCNLP}, pages 6251--6256.

\bibitem[{Hamaguchi et~al.(2017)Hamaguchi, Oiwa, Shimbo, and
  Matsumoto}]{hamaguchi2017knowledge}
Takuo Hamaguchi, Hidekazu Oiwa, Masashi Shimbo, and Yuji Matsumoto. 2017.
\newblock \href {https://www.ijcai.org/Proceedings/2017/0250.pdf} {{Knowledge
  Transfer for Out-of-Knowledge-Base Entities: A Graph Neural Network
  Approach}}.
\newblock In \emph{Proceedings of IJCAI}, pages 1802--1808.

\bibitem[{Hamilton et~al.(2017)Hamilton, Ying, and
  Leskovec}]{hamilton2017inductive}
William~L. Hamilton, Rex Ying, and Jure Leskovec. 2017.
\newblock \href
  {https://papers.nips.cc/paper/6703-inductive-representation-learning-on-large-graphs.pdf}
  {{Inductive Representation Learning on Large Graphs}}.
\newblock In \emph{Advances in Neural Information Processing Systems (NIPS)},
  pages 1025--1035.

\bibitem[{Han et~al.(2018{\natexlab{a}})Han, Liu, and Sun}]{han2018neural}
Xu~Han, Zhiyuan Liu, and Maosong Sun. 2018{\natexlab{a}}.
\newblock \href
  {https://www.aaai.org/ocs/index.php/AAAI/AAAI18/paper/view/16691} {{Neural
  Knowledge Acquisition via Mutual Attention Between Knowledge Graph and
  Text}}.
\newblock In \emph{Proceedings of AAAI}, pages 4832--4839.

\bibitem[{Han et~al.(2018{\natexlab{b}})Han, Zhu, Yu, Wang, Yao, Liu, and
  Sun}]{han-etal-2018-fewrel}
Xu~Han, Hao Zhu, Pengfei Yu, Ziyun Wang, Yuan Yao, Zhiyuan Liu, and Maosong
  Sun. 2018{\natexlab{b}}.
\newblock \href {https://doi.org/10.18653/v1/D18-1514} {{F}ew{R}el: {A
  Large-Scale Supervised Few-Shot Relation Classification Dataset with
  State-of-the-Art Evaluation}}.
\newblock In \emph{Proceedings of EMNLP}, pages 4803--4809.

\bibitem[{Hayashi et~al.(2020)Hayashi, Hu, Xiong, and
  Neubig}]{hayashi2019latent}
Hiroaki Hayashi, Zecong Hu, Chenyan Xiong, and Graham Neubig. 2020.
\newblock \href {https://aaai.org/ojs/index.php/AAAI/article/view/6298}
  {{Latent Relation Language Models}}.
\newblock In \emph{Proceedings of AAAI}, pages 7911--7918.

\bibitem[{Howard and Ruder(2018)}]{howard-ruder-2018-universal}
Jeremy Howard and Sebastian Ruder. 2018.
\newblock \href {https://doi.org/10.18653/v1/P18-1031} {{Universal Language
  Model Fine-tuning for Text Classification}}.
\newblock In \emph{Proceedings of ACL}, pages 328--339.

\bibitem[{Kassner and Sch{\"u}tze(2020)}]{kassner-schutze-2020-negated}
Nora Kassner and Hinrich Sch{\"u}tze. 2020.
\newblock \href {https://www.aclweb.org/anthology/2020.acl-main.698} {{Negated
  and Misprimed Probes for Pretrained Language Models: Birds Can Talk, But
  Cannot Fly}}.
\newblock In \emph{Proceedings of ACL}, pages 7811--7818.

\bibitem[{Kazemi and Poole(2018)}]{kazemi2018simple}
Seyed~Mehran Kazemi and David Poole. 2018.
\newblock \href
  {https://proceedings.neurips.cc/paper/2018/file/b2ab001909a8a6f04b51920306046ce5-Paper.pdf}
  {{SimplE Embedding for Link Prediction in Knowledge Graphs}}.
\newblock In \emph{Advances in Neural Information Processing Systems
  (NeurIPS)}, pages 4284--4295.

\bibitem[{Lan et~al.(2020)Lan, Chen, Goodman, Gimpel, Sharma, and
  Soricut}]{lan2019albert}
Zhenzhong Lan, Mingda Chen, Sebastian Goodman, Kevin Gimpel, Piyush Sharma, and
  Radu Soricut. 2020.
\newblock \href {https://arxiv.org/pdf/1909.11942.pdf} {{ALBERT: A Lite BERT
  for Self-supervised Learning of Language Representations}}.
\newblock In \emph{Proceedings of ICLR}.

\bibitem[{Lin et~al.(2015)Lin, Liu, Sun, Liu, and Zhu}]{lin2015learning}
Yankai Lin, Zhiyuan Liu, Maosong Sun, Yang Liu, and Xuan Zhu. 2015.
\newblock \href
  {https://www.aaai.org/ocs/index.php/AAAI/AAAI15/paper/viewFile/9571/9523/}
  {{Learning Entity and Relation Embeddings for Knowledge Graph Completion}}.
\newblock In \emph{Proceedings of AAAI}, pages 2181--2187.

\bibitem[{Liu et~al.(2019{\natexlab{a}})Liu, Gardner, Belinkov, Peters, and
  Smith}]{liu-etal-2019-linguistic}
Nelson~F. Liu, Matt Gardner, Yonatan Belinkov, Matthew~E. Peters, and Noah~A.
  Smith. 2019{\natexlab{a}}.
\newblock \href {https://doi.org/10.18653/v1/N19-1112} {{Linguistic Knowledge
  and Transferability of Contextual Representations}}.
\newblock In \emph{Proceedings of NAACL-HLT}, pages 1073--1094.

\bibitem[{Liu et~al.(2020)Liu, Zhou, Zhao, Wang, Ju, Deng, and Wang}]{liu2019k}
Weijie Liu, Peng Zhou, Zhe Zhao, Zhiruo Wang, Qi~Ju, Haotang Deng, and Ping
  Wang. 2020.
\newblock \href {https://doi.org/10.1609/aaai.v34i03.5681} {{K-BERT:} {Enabling
  Language Representation with Knowledge Graph}}.
\newblock In \emph{Proceedings of AAAI}, pages 2901--2908.

\bibitem[{Liu et~al.(2019{\natexlab{b}})Liu, He, Chen, and
  Gao}]{liu-etal-2019-multi}
Xiaodong Liu, Pengcheng He, Weizhu Chen, and Jianfeng Gao. 2019{\natexlab{b}}.
\newblock \href {https://doi.org/10.18653/v1/P19-1441} {{Multi-Task Deep Neural
  Networks for Natural Language Understanding}}.
\newblock In \emph{Proceedings of ACL}, pages 4487--4496.

\bibitem[{Liu et~al.(2019{\natexlab{c}})Liu, Ott, Goyal, Du, Joshi, Chen, Levy,
  Lewis, Zettlemoyer, and Stoyanov}]{liu2019roberta}
Yinhan Liu, Myle Ott, Naman Goyal, Jingfei Du, Mandar Joshi, Danqi Chen, Omer
  Levy, Mike Lewis, Luke Zettlemoyer, and Veselin Stoyanov. 2019{\natexlab{c}}.
\newblock \href {https://arxiv.org/pdf/1907.11692v1.pdf} {{RoBERTa: A Robustly
  Optimized BERT Pretraining Approach}}.
\newblock \emph{CoRR}, cs.CL/1907.11692v1.

\bibitem[{Logan et~al.(2019)Logan, Liu, Peters, Gardner, and
  Singh}]{logan-etal-2019-baracks}
Robert Logan, Nelson~F. Liu, Matthew~E. Peters, Matt Gardner, and Sameer Singh.
  2019.
\newblock \href {https://doi.org/10.18653/v1/P19-1598} {{B}arack{'}s {Wife
  Hillary: Using Knowledge Graphs for Fact-Aware Language Modeling}}.
\newblock In \emph{Proceedings of ACL}, pages 5962--5971.

\bibitem[{Logeswaran et~al.(2019)Logeswaran, Chang, Lee, Toutanova, Devlin, and
  Lee}]{logeswaran-etal-2019-zero}
Lajanugen Logeswaran, Ming-Wei Chang, Kenton Lee, Kristina Toutanova, Jacob
  Devlin, and Honglak Lee. 2019.
\newblock \href {https://doi.org/10.18653/v1/P19-1335} {{Zero-Shot Entity
  Linking by Reading Entity Descriptions}}.
\newblock In \emph{Proceedings of ACL}, pages 3449--3460.

\bibitem[{McCloskey and Cohen(1989)}]{mccloskey1989catastrophic}
Michael McCloskey and Neal~J Cohen. 1989.
\newblock \href
  {https://www.sciencedirect.com/science/article/pii/S0079742108605368}
  {Catastrophic interference in connectionist networks: {T}he sequential
  learning problem}.
\newblock In \emph{Psychology of learning and motivation}, volume~24, pages
  109--165. Elsevier.

\bibitem[{Mikolov et~al.(2013)Mikolov, Sutskever, Chen, Corrado, and
  Dean}]{mikolov2013distributed}
Tomas Mikolov, Ilya Sutskever, Kai Chen, Gregory~S. Corrado, and Jeffrey Dean.
  2013.
\newblock \href
  {http://papers.nips.cc/paper/5021-distributed-representations-of-words-and-phrases-and-their-compositionality}
  {{Distributed Representations of Words and Phrases and their
  Compositionality}}.
\newblock In \emph{Advances in Neural Information Processing Systems (NIPS)},
  pages 3111--3119.

\bibitem[{Miller(1995)}]{miller1995wordnet}
George~A. Miller. 1995.
\newblock \href {https://doi.org/10.1145/219717.219748} {{WordNet: A Lexical
  Database for English}}.
\newblock \emph{Commun. ACM}, 38(11):39--41.

\bibitem[{Ott et~al.(2019)Ott, Edunov, Baevski, Fan, Gross, Ng, Grangier, and
  Auli}]{ott2019fairseq}
Myle Ott, Sergey Edunov, Alexei Baevski, Angela Fan, Sam Gross, Nathan Ng,
  David Grangier, and Michael Auli. 2019.
\newblock \href {https://doi.org/10.18653/v1/N19-4009} {fairseq: {A Fast,
  Extensible Toolkit for Sequence Modeling}}.
\newblock In \emph{Proceedings of NAACL-HLT (Demonstrations)}, pages 48--53.

\bibitem[{Pennington et~al.(2014)Pennington, Socher, and
  Manning}]{pennington-etal-2014-glove}
Jeffrey Pennington, Richard Socher, and Christopher Manning. 2014.
\newblock \href {https://doi.org/10.3115/v1/D14-1162} {{G}lo{V}e: {Global
  Vectors for Word Representation}}.
\newblock In \emph{Proceedings of EMNLP}, pages 1532--1543.

\bibitem[{Peters et~al.(2018)Peters, Neumann, Iyyer, Gardner, Clark, Lee, and
  Zettlemoyer}]{peters-etal-2018-deep}
Matthew Peters, Mark Neumann, Mohit Iyyer, Matt Gardner, Christopher Clark,
  Kenton Lee, and Luke Zettlemoyer. 2018.
\newblock \href {https://doi.org/10.18653/v1/N18-1202} {{Deep Contextualized
  Word Representations}}.
\newblock In \emph{Proceedings of NAACL-HLT}, pages 2227--2237.

\bibitem[{Peters et~al.(2019)Peters, Neumann, Logan, Schwartz, Joshi, Singh,
  and Smith}]{peters-etal-2019-knowledge}
Matthew~E. Peters, Mark Neumann, Robert Logan, Roy Schwartz, Vidur Joshi,
  Sameer Singh, and Noah~A. Smith. 2019.
\newblock \href {https://doi.org/10.18653/v1/D19-1005} {{Knowledge Enhanced
  Contextual Word Representations}}.
\newblock In \emph{Proceedings of EMNLP-IJCNLP}, pages 43--54.

\bibitem[{Petroni et~al.(2019)Petroni, Rockt{\"a}schel, Riedel, Lewis, Bakhtin,
  Wu, and Miller}]{petroni2019language}
Fabio Petroni, Tim Rockt{\"a}schel, Sebastian Riedel, Patrick Lewis, Anton
  Bakhtin, Yuxiang Wu, and Alexander Miller. 2019.
\newblock \href {https://doi.org/10.18653/v1/D19-1250} {{Language Models as
  Knowledge Bases}?}
\newblock In \emph{Proceedings of EMNLP-IJCNLP}, pages 2463--2473.

\bibitem[{Poerner et~al.(2020)Poerner, Waltinger, and
  Sch{\"u}tze}]{Poerner2019EBERT}
Nina Poerner, Ulli Waltinger, and Hinrich Sch{\"u}tze. 2020.
\newblock \href {https://www.aclweb.org/anthology/2020.findings-emnlp.71}
  {{E}-{BERT}: {Efficient-Yet-Effective Entity Embeddings} for {BERT}}.
\newblock In \emph{Findings of the Association for Computational Linguistics:
  EMNLP 2020}, pages 803--818.

\bibitem[{Radford et~al.(2018)Radford, Narasimhan, Salimans, and
  Sutskever}]{radford2018improving}
Alec Radford, Karthik Narasimhan, Tim Salimans, and Ilya Sutskever. 2018.
\newblock \href
  {https://s3-us-west-2.amazonaws.com/openai-assets/research-covers/language-unsupervised/language_understanding_paper.pdf}
  {{Improving Language Understanding by Generative Pre-Training}}.
\newblock In \emph{Technical report, OpenAI}.

\bibitem[{Raffel et~al.(2020)Raffel, Shazeer, Roberts, Lee, Narang, Matena,
  Zhou, Li, and Liu}]{raffel2019exploring}
Colin Raffel, Noam Shazeer, Adam Roberts, Katherine Lee, Sharan Narang, Michael
  Matena, Yanqi Zhou, Wei Li, and Peter~J. Liu. 2020.
\newblock \href {http://jmlr.org/papers/v21/20-074.html} {{Exploring the Limits
  of Transfer Learning with a Unified Text-to-Text Transformer}}.
\newblock \emph{Journal of Machine Learning Research}, 21(140):1--67.

\bibitem[{Sennrich et~al.(2016)Sennrich, Haddow, and
  Birch}]{sennrich-etal-2016-neural}
Rico Sennrich, Barry Haddow, and Alexandra Birch. 2016.
\newblock \href {https://doi.org/10.18653/v1/P16-1162} {{Neural Machine
  Translation of Rare Words with Subword Units}}.
\newblock In \emph{Proceedings of ACL}, pages 1715--1725.

\bibitem[{Shi and Weninger(2018)}]{shi2018open}
Baoxu Shi and Tim Weninger. 2018.
\newblock \href
  {https://www.aaai.org/ocs/index.php/AAAI/AAAI18/paper/view/16055}
  {{Open-World Knowledge Graph Completion}}.
\newblock In \emph{Proceedings of AAAI}, pages 1957--1964.

\bibitem[{Snell et~al.(2017)Snell, Swersky, and Zemel}]{snell2017prototypical}
Jake Snell, Kevin Swersky, and Richard Zemel. 2017.
\newblock \href
  {https://proceedings.neurips.cc/paper/2017/file/cb8da6767461f2812ae4290eac7cbc42-Paper.pdf}
  {{Prototypical Networks for Few-shot Learning}}.
\newblock In \emph{Advances in Neural Information Processing Systems (NIPS)},
  pages 4077--4087.

\bibitem[{Sun et~al.(2019)Sun, Deng, Nie, and Tang}]{sun2019rotate}
Zhiqing Sun, Zhi-Hong Deng, Jian-Yun Nie, and Jian Tang. 2019.
\newblock \href {https://openreview.net/pdf?id=HkgEQnRqYQ} {{RotatE: Knowledge
  Graph Embedding by Relational Rotation in Complex Space}}.
\newblock In \emph{Proceedings of ICLR}.

\bibitem[{Trouillon et~al.(2016)Trouillon, Welbl, Riedel, Gaussier, and
  Bouchard}]{trouillon2016complex}
Th\'{e}o Trouillon, Johannes Welbl, Sebastian Riedel, \'{E}ric Gaussier, and
  Guillaume Bouchard. 2016.
\newblock \href {http://proceedings.mlr.press/v48/trouillon16.pdf} {{Complex
  Embeddings for Simple Link Prediction}}.
\newblock In \emph{Proceedings of ICML}, pages 2071--2080.

\bibitem[{Vaswani et~al.(2017)Vaswani, Shazeer, Parmar, Uszkoreit, Jones,
  Gomez, Kaiser, and Polosukhin}]{vaswani2017attention}
Ashish Vaswani, Noam Shazeer, Niki Parmar, Jakob Uszkoreit, Llion Jones,
  Aidan~N Gomez, {\L}ukasz Kaiser, and Illia Polosukhin. 2017.
\newblock \href
  {https://papers.nips.cc/paper/7181-attention-is-all-you-need.pdf} {{Attention
  is All you Need}}.
\newblock In \emph{Advances in Neural Information Processing Systems (NIPS)},
  pages 5998--6008.

\bibitem[{Wang et~al.(2019{\natexlab{a}})Wang, Hula, Xia, Pappagari, McCoy,
  Patel, Kim, Tenney, Huang, Yu, Jin, Chen, Van~Durme, Grave, Pavlick, and
  Bowman}]{wang-etal-2019-tell}
Alex Wang, Jan Hula, Patrick Xia, Raghavendra Pappagari, R.~Thomas McCoy, Roma
  Patel, Najoung Kim, Ian Tenney, Yinghui Huang, Katherin Yu, Shuning Jin,
  Berlin Chen, Benjamin Van~Durme, Edouard Grave, Ellie Pavlick, and Samuel~R.
  Bowman. 2019{\natexlab{a}}.
\newblock \href {https://doi.org/10.18653/v1/P19-1439} {{Can You Tell Me How to
  Get Past Sesame Street? Sentence-Level Pretraining Beyond Language
  Modeling}}.
\newblock In \emph{Proceedings of ACL}, pages 4465--4476.

\bibitem[{Wang et~al.(2019{\natexlab{b}})Wang, Singh, Michael, Hill, Levy, and
  Bowman}]{wang-etal-2018-glue}
Alex Wang, Amanpreet Singh, Julian Michael, Felix Hill, Omer Levy, and Samuel
  Bowman. 2019{\natexlab{b}}.
\newblock \href {https://openreview.net/pdf?id=rJ4km2R5t7} {{GLUE}: {A
  Multi-Task Benchmark and Analysis Platform for Natural Language
  Understanding}}.
\newblock In \emph{Proceedings of ICLR}.

\bibitem[{Wang et~al.(2019{\natexlab{c}})Wang, Han, Li, and
  Pan}]{Wang2019LogicAB}
PeiFeng Wang, Jialong Han, Chenliang Li, and Rong Pan. 2019{\natexlab{c}}.
\newblock \href {https://doi.org/10.1609/aaai.v33i01.33017152} {Logic attention
  based neighborhood aggregation for inductive knowledge graph embedding}.
\newblock In \emph{Proceedings of AAAI}, pages 7152--7159.

\bibitem[{Wang et~al.(2020)Wang, Tang, Duan, Wei, Huang, Ji, Cao, Jiang, and
  Zhou}]{Wang2020KAdapter}
Ruize Wang, Duyu Tang, Nan Duan, Zhongyu Wei, Xuanjing Huang, Jianshu Ji,
  Cuihong Cao, Daxin Jiang, and Ming Zhou. 2020.
\newblock \href {https://arxiv.org/abs/2002.01808v3} {{K-Adapter: Infusing
  Knowledge into Pre-Trained Models with Adapters}}.
\newblock \emph{CoRR}, cs.CL/2002.01808v3.

\bibitem[{Wang et~al.(2014)Wang, Zhang, Feng, and Chen}]{wang2014knowledge}
Zhen Wang, Jianwen Zhang, Jianlin Feng, and Zheng Chen. 2014.
\newblock \href {https://doi.org/10.3115/v1/D14-1167} {{Knowledge Graph and
  Text Jointly Embedding}}.
\newblock In \emph{Proceedings of EMNLP}, pages 1591--1601.

\bibitem[{Williams et~al.(2018)Williams, Nangia, and Bowman}]{N18-1101}
Adina Williams, Nikita Nangia, and Samuel Bowman. 2018.
\newblock \href {http://aclweb.org/anthology/N18-1101} {{A Broad-Coverage
  Challenge Corpus for Sentence Understanding through Inference}}.
\newblock In \emph{Proceedings of NAACL-HLT}, pages 1112--1122.

\bibitem[{Wu et~al.(2020)Wu, Petroni, Josifoski, Riedel, and
  Zettlemoyer}]{wu2019zero}
Ledell Wu, Fabio Petroni, Martin Josifoski, Sebastian Riedel, and Luke
  Zettlemoyer. 2020.
\newblock \href {https://www.aclweb.org/anthology/2020.emnlp-main.519}
  {Scalable {Z}ero-shot {E}ntity {L}inking with {D}ense {E}ntity {R}etrieval}.
\newblock In \emph{Proceedings of EMNLP}, pages 6397--6407.

\bibitem[{Xie et~al.(2016)Xie, Liu, Jia, Luan, and
  Sun}]{Xie:2016:RLK:3016100.3016273}
Ruobing Xie, Zhiyuan Liu, Jia Jia, Huanbo Luan, and Maosong Sun. 2016.
\newblock \href
  {http://www.aaai.org/ocs/index.php/AAAI/AAAI16/paper/view/12216}
  {{Representation Learning of Knowledge Graphs with Entity Descriptions}}.
\newblock In \emph{Proceedings of AAAI}, pages 2659--2665.

\bibitem[{Xiong et~al.(2019)Xiong, Du, Wang, and Veselin}]{xiong2019pretrained}
Wenhan Xiong, Jingfei Du, William~Yang Wang, and Stoyanov Veselin. 2019.
\newblock \href {https://openreview.net/pdf?id=BJlzm64tDH} {{Pretrained
  Encyclopedia: Weakly Supervised Knowledge-Pretrained Language Model}}.
\newblock In \emph{Proceedings of ICLR}.

\bibitem[{Yamada et~al.(2016)Yamada, Shindo, Takeda, and
  Takefuji}]{yamada2016joint}
Ikuya Yamada, Hiroyuki Shindo, Hideaki Takeda, and Yoshiyasu Takefuji. 2016.
\newblock \href {https://doi.org/10.18653/v1/K16-1025} {{Joint Learning of the
  Embedding of Words and Entities for Named Entity Disambiguation}}.
\newblock In \emph{Proceedings of CoNLL}, pages 250--259.

\bibitem[{Yang and Mitchell(2017)}]{yang-mitchell-2017-leveraging}
Bishan Yang and Tom Mitchell. 2017.
\newblock \href {https://doi.org/10.18653/v1/P17-1132} {{Leveraging Knowledge
  Bases in {LSTM}s for Improving Machine Reading}}.
\newblock In \emph{Proceedings of ACL}, pages 1436--1446.

\bibitem[{Yang et~al.(2015)Yang, Yih, He, Gao, and Deng}]{yang2015embedding}
Bishan Yang, Scott Wen-tau Yih, Xiaodong He, Jianfeng Gao, and Li~Deng. 2015.
\newblock \href
  {https://www.microsoft.com/en-us/research/publication/embedding-entities-and-relations-for-learning-and-inference-in-knowledge-bases/}
  {{Embedding Entities and Relations for Learning and Inference in Knowledge
  Bases}}.
\newblock In \emph{Proceedings of ICLR}.

\bibitem[{Yang et~al.(2019)Yang, Dai, Yang, Carbonell, Salakhutdinov, and
  Le}]{yang2019xlnet}
Zhilin Yang, Zihang Dai, Yiming Yang, Jaime~G. Carbonell, Ruslan Salakhutdinov,
  and Quoc~V. Le. 2019.
\newblock \href
  {http://papers.nips.cc/paper/8812-xlnet-generalized-autoregressive-pretraining-for-language-understanding}
  {{XLNet}: {Generalized Autoregressive Pretraining for Language
  Understanding}}.
\newblock In \emph{Advances in Neural Information Processing Systems
  (NeurIPS)}, pages 5754--5764.

\bibitem[{Zaremoodi et~al.(2018)Zaremoodi, Buntine, and
  Haffari}]{zaremoodi-etal-2018-adaptive}
Poorya Zaremoodi, Wray Buntine, and Gholamreza Haffari. 2018.
\newblock \href {https://doi.org/10.18653/v1/P18-2104} {{Adaptive Knowledge
  Sharing in Multi-Task Learning: Improving Low-Resource Neural Machine
  Translation}}.
\newblock In \emph{Proceedings of ACL}, pages 656--661.

\bibitem[{Zhang et~al.(2017)Zhang, Zhong, Chen, Angeli, and
  Manning}]{zhang-etal-2017-position}
Yuhao Zhang, Victor Zhong, Danqi Chen, Gabor Angeli, and Christopher~D.
  Manning. 2017.
\newblock \href {https://doi.org/10.18653/v1/D17-1004} {{Position-aware
  Attention and Supervised Data Improve Slot Filling}}.
\newblock In \emph{Proceedings of EMNLP}, pages 35--45.

\bibitem[{Zhang et~al.(2019)Zhang, Han, Liu, Jiang, Sun, and
  Liu}]{zhang-etal-2019-ernie}
Zhengyan Zhang, Xu~Han, Zhiyuan Liu, Xin Jiang, Maosong Sun, and Qun Liu. 2019.
\newblock \href {https://doi.org/10.18653/v1/P19-1139} {{ERNIE}: {Enhanced
  Language Representation with Informative Entities}}.
\newblock In \emph{Proceedings of ACL}, pages 1441--1451.

\bibitem[{Zhu et~al.(2015)Zhu, Kiros, Zemel, Salakhutdinov, Urtasun, Torralba,
  and Fidler}]{zhu2015aligning}
Yukun Zhu, Ryan Kiros, Rich Zemel, Ruslan Salakhutdinov, Raquel Urtasun,
  Antonio Torralba, and Sanja Fidler. 2015.
\newblock \href {https://doi.org/10.1109/ICCV.2015.11} {{Aligning Books and
  Movies: Towards Story-Like Visual Explanations by Watching Movies and Reading
  Books}}.
\newblock In \emph{Proceedings of ICCV}, pages 19--27.

\bibitem[{Zhu et~al.(2019)Zhu, Xu, Tang, and Qu}]{zhu2019graphvite}
Zhaocheng Zhu, Shizhen Xu, Jian Tang, and Meng Qu. 2019.
\newblock \href {https://doi.org/10.1145/3308558.3313508} {{GraphVite}: {A}
  {H}igh-{P}erformance {CPU-GPU} {H}ybrid {S}ystem for {N}ode {E}mbedding}.
\newblock In \emph{Proceedings of WWW}, pages 2494--2504.

\end{thebibliography}

\end{document}